\documentclass[preprint,12pt]{elsarticle}

\usepackage{dsfont}
\usepackage{algorithm}
\usepackage{algpseudocode}
\usepackage{setspace}
\usepackage{graphics}
\usepackage{graphicx}
\usepackage{epsfig}
\usepackage{epstopdf}
\usepackage{amssymb}
\usepackage{amsthm}
\usepackage{amsmath}
\usepackage{mathtools}
\usepackage{bm}
\usepackage{geometry}
\usepackage{subfigure}
\usepackage{url}
\PassOptionsToPackage{hyphens}{url}
\usepackage{etex}
\usepackage[table]{xcolor}
\usepackage{array}
\usepackage{multirow} 
\biboptions{round,authoryear}

\DeclareMathOperator*{\argmin}{arg\,min}

\journal{}

\begin{document}
\begin{frontmatter}



\title{Estimation of Fiber Orientations Using Neighborhood Information}


\author[label1]{Chuyang Ye\corref{cor1}}
\author[label2]{Jiachen Zhuo}
\author[label2]{Rao P. Gullapalli}
\author[label3]{Jerry L. Prince}

\address[label1]{Brainnetome Center, Institute of Automation, Chinese Academy of Sciences, Beijing, China}
\address[label2]{Department of Radiology, University of Maryland School of Medicine,\\Baltimore, MD, USA}
\address[label3]{Department of Electrical and Computer Engineering,\\Johns Hopkins University, Baltimore, MD, USA}
\cortext[cor1]{Address: Intelligence Building 504, 95 Zhongguancun East Road, 
Beijing, China, 100190. 
\\ \indent Email address: \texttt{chuyang.ye@nlpr.ia.ac.cn}}

\begin{abstract}
Data from \textit{diffusion magnetic resonance imaging} (dMRI) can be used to reconstruct fiber tracts, for example, in muscle and white
matter. Estimation of \textit{fiber orientations} (FOs)
is a crucial step in the reconstruction process and these estimates can be corrupted by noise.
In this paper, a new method called \textit{Fiber Orientation Reconstruction using Neighborhood Information} (FORNI)
is described and shown to reduce the effects of noise and improve FO estimation performance
by incorporating spatial consistency. 
FORNI uses a fixed tensor basis to model the diffusion weighted signals,
which has the advantage of providing an explicit relationship between the basis
vectors and the FOs.  FO spatial coherence is encouraged using weighted
$\ell_{1}$-norm regularization terms, which contain the interaction
of directional information between neighbor voxels.  Data fidelity
is encouraged using a squared error between the observed and reconstructed
diffusion weighted signals.  After appropriate weighting of these
competing objectives, the resulting objective function is
minimized using a block coordinate descent algorithm, and a
straightforward parallelization strategy is used to speed up
processing.  Experiments were performed on a digital crossing
phantom, \textit{ex vivo} tongue dMRI data, and \textit{in vivo}
brain dMRI data for both qualitative and quantitative evaluation. The
results demonstrate that FORNI improves the quality of
FO estimation over other state of the art algorithms. 
\end{abstract}


\end{frontmatter}


\section{Introduction}
\label{sec:intro}

By capturing both the magnitude and the anisotropy of water
diffusion, \textit{diffusion magnetic resonance imaging} (dMRI) provides a noninvasive
means to reconstruct fiber tracts, for example, in white matter and
muscle~\citep{johansen}. \textit{Diffusion tensor imaging} (DTI), 
which is a basic dMRI strategy, models the water diffusion using a
symmetric positive definite tensor~\citep{basser}. Since DTI is known
to be insufficient to represent crossing fiber tracts, more advanced
dMRI techniques, such as \textit{high angular resolution diffusion imaging}
(HARDI)~\citep{HARDI} and \textit{diffusion spectrum imaging}
(DSI)~\citep{DSI}, have been developed. 

In order to carry out tractography~\citep{mori,basser2,qazi,reisert}
and volumetric fiber tract segmentation~\citep{bazin,nazem,yendiki,NEIN}, \textit{fiber
orientations} (FOs) are computed from the dMRI data. In tractography, fiber streamlines are propagated according to the computed FOs or the distribution of FOs, and in volumetric tract segmentation the FO is a key feature upon which the voxels are labeled. Since accurate estimation
of FOs is critical in these algorithms, it has been a major
topic of research. For example, spherical
deconvolution~\citep{tournier,tournier2,cheng2,jeurissen}, $q$-ball
reconstruction~\citep{qball,hess,qbi}, multi-tensor
models~\citep{landman,peled,behrens,ramirez,zhou,liu,BAMBI}, and ensemble average propagator methods~\citep{michailovich,rathi,wedeen2008,pickalov,ozarslan,merlet2013} have been developed so that multiple FOs can be estimated in each voxel.

A large number of diffusion gradient directions may be required to accurately estimate FOs when fiber tracts cross, which takes a long acquisition time and
limits the use of dMRI in clinical practice~\citep{bilgic}. Therefore,
methods have been developed to reduce the required number of gradient
directions so that the dMRI acquisition is clinically achievable.
Because the number of crossing FOs in a voxel is small, modeling the
diffusion data as having arisen from a sparse subset of basis sources
and solving the resulting optimization problem using sparsity
regularization is particularly effective~\citep{ramirez, landman, 
daducci, merlet, zhou, michailovich, rathi}. The basis has been
selected to be prolate diffusion
tensors~\citep{ramirez, landman, daducci, zhou}, spherical
ridgelets~\citep{michailovich, rathi}, and spherical polar Fourier
basis~\citep{merlet}.

\begin{figure}[t]
  \centering
	\subfigure[CFARI]{
		\includegraphics[width=0.25\columnwidth]{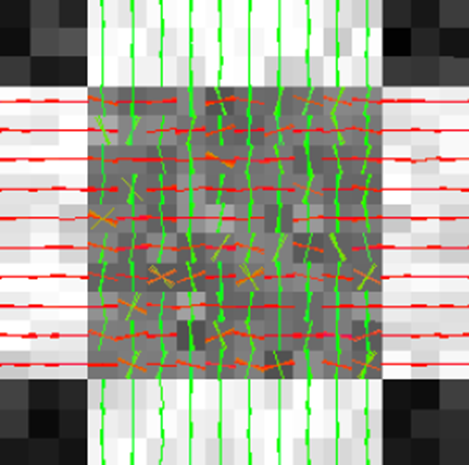}
		\label{fig:toy_cfari}
	}
  \subfigure[FORNI]{
		\includegraphics[width=0.25\columnwidth]{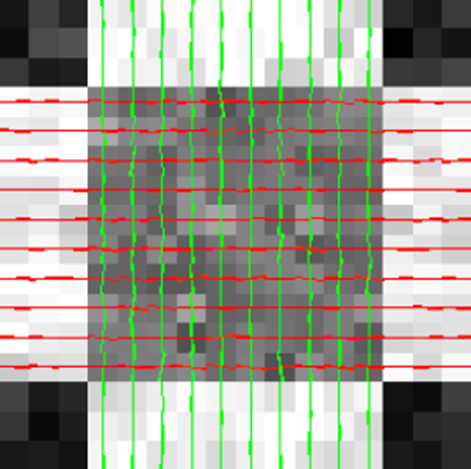}
		\label{fig:toy_foruni}
	}
\caption{A 3D toy example of FO estimation on two simulated crossing
  tracts in the axial view (the $x$-$y$ plane): (a) voxelwise FO estimation using the CFARI
  algorithm~\citep{landman} and (b) FOs estimated by the proposed
  method (FORNI) incorporating spatial coherence of FOs.} 
\label{fig:toy}
\end{figure}

Noise can have a deleterious effect on FO estimation, especially in areas where
fibers cross~\citep{cheng,aranda}.
A 3D toy example of two crossing tracts is shown in the axial view (the $x$-$y$ plane) in Figure~\ref{fig:toy_cfari}, where noise is added to the simulated dMRI data. Here, the CFARI
algorithm~\citep{landman} estimates FOs at each voxel independently;
it yields noisy estimates and occasionally fails to yield a second
direction at all (Figure~\ref{fig:toy_cfari}). 

To reduce the effect of noise, spatial coherence (or smoothness) has been used to improve FO estimation. 
In~\citet{becker} diffusion weighted images are smoothed
before FO estimation.  In~\citet{sigurdsson}, FOs are smoothed after
voxelwise estimation using the CFARI method. 
In~\citet{duits}, FOs are smoothed using left-invariant diffusions on the space of positions and orientations.
\citet{tournier2013} and~\citet{reisert2011} incorporate the continuity of FOs as regularization terms in the estimation to enforce FO smoothness, but sparsity regularization is not used.
There are also methods that seek to simultaneously estimate and smooth FOs by
combining spatial continuity with sparsity.
In~\citet{michailovich} and \citet{rathi}, the TV-norm of diffusion
weighted images is incorporated as a smoothness regularization term in
the objective function.  In~\citet{ramirez} and \citet{zhou}, spatial
consistency of FOs is encouraged by adding regularization terms that smooth the mixture fractions of each basis tensor. However, spatial coherence is preserved in an indirect way in \citet{michailovich},
\citet{rathi}, \citet{ramirez}, and~\citet{zhou} because the
objective functions do not explicitly model and smooth the directional information
in the FOs. Recently, \citet{auria} define a spatially structured sparsity regularization term to incorporate directional information in the sparse reconstruction of FOs.

In this paper, we present the method \textit{Fiber Orientation Reconstruction
using Neighborhood Information} (FORNI), which is an FO estimation
algorithm that incorporates spatial coherence. 
Preliminary results of this work were presented in a conference paper~\citep{CDMRI}.
An example of the FORNI FO estimation on the toy example in Figure~\ref{fig:toy_cfari} is shown in Figure~\ref{fig:toy_foruni}.
In contrast to most previous works, we form an objective function that directly encodes the directional information in the neighborhood to encourage spatial coherence of FOs.
Specifically, a fixed tensor basis is used to represent diffusion weighted signals, which has the advantage of providing an explicit relationship between the basis and FOs.
Spatial coherence is encouraged using weighted $\ell_{1}$-norm regularization, where the
interaction of directional information in neighboring voxels is
modeled. In the weighted $\ell_{1}$-norm regularization terms, basis
directions that are more consistent with the FOs in the neighborhood
are encouraged.  Data fidelity is encouraged using a term that measures agreement between the observed and reconstructed diffusion signals. The resulting objective
function is minimized using a block coordinate descent algorithm, and
a straightforward parallelization strategy is used to speed up processing.

The remainder of the paper is organized as
follows. Section~\ref{sec:methods} describes the FORNI algorithm and Section~\ref{sec:exp} presents the experiments on a
digital crossing phantom, \textit{ex vivo} tongue dMRI data, and
\textit{in vivo} brain dMRI data for qualitative and quantitative
evaluation. Section~\ref{sec:discussion} discusses the results and
future works. Finally, Section~\ref{sec:conclusion} concludes the
paper.

\section{Methods}
\label{sec:methods}

In this section, we first provide a background on diffusion signal
modeling using a fixed tensor basis. Then, we describe our approach to
the incorporation of directional information from neighboring voxels to
improve FO estimation. Finally, the resulting objective function and the
optimization strategy are presented. 

\subsection{A Multi-tensor Model with a Fixed Tensor Basis}

Using the unified framework presented in~\citet{jian}, the diffusion
weighted signal at each voxel can be modeled as  
\begin{eqnarray}
S({\bm{q}}) = S_0\int\limits_{\mathcal{M}}{f(x)R(\bm{q},x)}\mathrm{d}x,
\label{eqn:dwi_unified}
\end{eqnarray}
where $x$ is a point on a smooth manifold $\mathcal{M}$, $S(\bm{q})$
is the diffusion weighted signal with the diffusion gradient $\bm{q}$,
$S_{0}$ is the baseline signal without diffusion weighting, $f(x)$ is
a probability density function, and $R(\bm{q},x)$ is a kernel
function. 

The diffusion signals can be represented by a basis, and one commonly
used basis is a set of fixed prolate tensors~\citep{landman, zhou, ramirez, daducci}.  The \textit{primary eigenvector}~(PEV) of each basis tensor represents a possible FO and is referred to as a basis direction. In this work, we use the tensor
basis comprising $N=289$ prolate tensors $\textbf{D}_{i}$ whose PEVs
$\bm{v}_{i}$ are approximately evenly oriented over the unit sphere.
These basis directions were determined by tessellating an octahedron, and the number of basis directions ($N=289$) lies in the range of previously used numbers~\citep{ramirez,landman,auria}.
The shape of the basis tensor is determined by its eigenvalues ($\lambda_{1}\geq \lambda_{2} \geq \lambda_{3} > 0$). The second and third eigenvalues are set equal, and each eigenvalue is determined by
examining the diffusion tensors of a noncrossing fiber
tract~\citep{landman}.

With this tensor basis, we have $\mathcal{M}=\mathcal{S}^2$ (a unit
sphere), $x=\bm{v}$ (a unit vector),
$f(\bm{v})=f_{i}\delta(\bm{v};\bm{v}_{i})$, and
$R(\bm{q},\bm{v}_{i})=e^{-\bm{q}^{T}\textbf{D}_{i}\bm{q}}$ (based on
the Stejskal-Tanner tensor formulation~\citep{stejskal}). By
normalizing the diffusion gradient as
$\tilde{\bm{q}}=\bm{q}/|\bm{q}|$, the gradient direction
$\tilde{\bm{q}}$ is associated with a constant $b$ determined by the
imaging sequence. Taking image noise into account,
Eq.~(\ref{eqn:dwi_unified}) then becomes~\citep{landman} 
\begin{eqnarray}
S(\bm{q}) =
S_0\sum\limits_{i=1}^{N}f_{i}e^{-b\tilde{\bm{q}}^{T}\textbf{D}_{i}\tilde{\bm{q}}}
+ n(\bm{q}),  
\label{eqn:dwi}
\end{eqnarray}
where $f_{i}$ is the (unknown) nonnegative mixture fraction for
$\textbf{D}_{i}$, $\sum_{i=1}^{N} f_{i} = 1$, and $n(\bm{q})$ is
noise. 

After defining $y(\bm{q}) = S(\bm{q})/S_{0}$ and $\eta(\bm{q}) =
n(\bm{q})/S_{0}$ and letting $K$ be the number of diffusion
gradient measurements, Eq.~(\ref{eqn:dwi}) can be written as 
\begin{eqnarray}
\bm{y} = \textbf{G}\bm{f} + \bm{\eta},
\label{equ:linear}
\end{eqnarray}
where $\bm{y}=(y({\bm{q}_1}),y({\bm{q}_2}),...,y({\bm{q}_K}))^{T}$,
$\textbf{G}$ is a $K\times 
N$ matrix comprising the attenuation terms
$G_{ki}=e^{-b_{k}\tilde{\bm{q}}_{k}^{T}\textbf{D}_{i}\tilde{\bm{q}}_{k}}$,
$\bm{f}=(f_{1},f_{2},...,f_{N})^{T}$, and
$\bm{\eta}=(\eta({\bm{q}_1}),\eta({\bm{q}_2}),...,\eta({\bm{q}_K}))^{T}$. 
Because the number of FOs at each voxel is usually small with respect
to the number of diffusion gradients, the mixture fractions can be
estimated using the following sparse reconstruction formulation 
\begin{eqnarray}
\hat{\bm{f}} = \argmin\limits_{\bm{f}\geq \bm{0},
  ||\bm{f}||_{1}=1}||\textbf{G}\bm{f}-\bm{y}||_{2}^{2} + \beta
||\bm{f}||_{0}. 
\label{equ:sr}
\end{eqnarray}

To solve Eq.~(\ref{equ:sr}), the constraint of $\sum_{i=1}^{N}
f_{i}=1$ is relaxed~\citep{landman,ramirez} and then the 
$\ell_{0}$-norm is replaced by the $\ell_{1}$-norm, yielding the
following simpler problem, 
\begin{eqnarray}
\hat{\bm{f}} = \argmin\limits_{\bm{f}\geq
  \bm{0}}||\textbf{G}\bm{f}-\bm{y}||_{2}^{2} + \beta ||\bm{f}||_{1}\,.  
\label{equ:obj}
\end{eqnarray} 
After solving this, the estimated vector $\hat{\bm{f}}$ is normalized
so that its elements add to one. 
Note that in this paper we interpret small mixture fractions as components of
isotropic diffusion; therefore, FOs are given by those basis
directions whose mixture fractions are greater than a threshold $f_{\mathrm{th}}$.  
Accordingly, by focusing on the estimation of mixture fractions, we
are also estimating the FOs. 

\subsection{FO Estimation Using Neighborhood Information}

\begin{figure}[t]
\centering
   \includegraphics[width=0.9\linewidth]{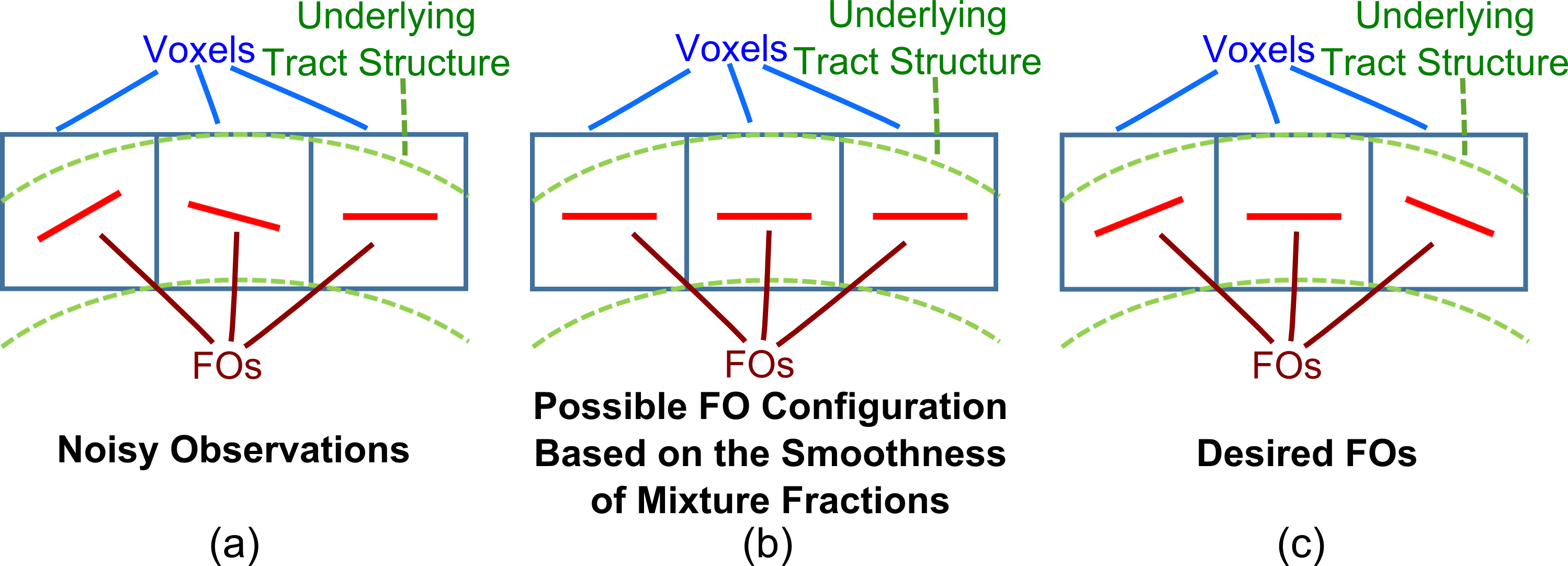}
\caption[]{A graphical example showing the potential difficulties when
  the smoothness of mixture fractions is used to represent spatial
  coherence of FOs.} 
\label{fig:graphical_eg}
\end{figure}

Incorporation of spatial coherence in the estimation of FOs can reduce
the effects of noise~\citep{michailovich}.  Some researchers have
incorporated neighborhood information in order to maintain smoothness
of mixture fractions~\citep{ramirez,zhou}.  But having smooth mixture
fractions does not equate to having smooth FO angles. For example,
suppose we have three side-by-side voxels $a$, $b$, and $c$, whose mixture
fractions are $\bm{f}_{a}=(1,0,...,0)^{T}$,
$\bm{f}_{b}=(0,1,0,...,0)^{T}$, and $\bm{f}_{c}=(0,0,1,0,...,0)^{T}$,
respectively. The magnitude of the difference
$||\bm{f}_{a}-\bm{f}_{b}||$ between the mixture fractions of $a$ and
$b$ is the same as the magnitude of the difference
$||\bm{f}_{a}-\bm{f}_{c}||$ between $a$ and $c$, while the desired
measure of difference should be related to the angles between the
basis directions with nonzero mixture fraction entries.
Figure~\ref{fig:graphical_eg} gives a graphical example of a
curved tract, where simply using the smoothness of mixture fractions
could lead to identical FOs along the tract instead of a desired
gradually-changing FO structure. This limitation exists in the case of
crossing fibers as well. In this work, we encourage spatial coherence
of FOs by explicitly incorporating the directional information in
neighboring voxels into the FO estimation. 
The symbols used in FORNI are listed in Table~\ref{tab:symbols}.

\begin{table}[t]
\centering
\caption{Symbols used in FORNI} 
\resizebox{0.65\columnwidth}{!}{
\begin{tabular}{l l}
\hline
\hline
Symbol & Definition\\
\hline
$\bm{v}_{i}$ & Basis direction\\
$i,q$ & Index for basis directions\\
$N$ & Number of basis directions\\
$\mathbf{G}$ & Dictionary matrix containing attenuation terms\\
$m,n,m_{0},m_{1}$ & Index for voxels \\
$M$ & Number of voxels\\
$w_{m,n}$ & Voxel similarity between $m$ and $n$\\
$\mathcal{N}_{m}$ & Neighborhood of $m$\\
$\mathbf{D}_{m}$ & Diffusion tensor at $m$ \\
$\bm{y}_{m}$ & Diffusion signal at $m$\\
$\bm{f}_{m}$ & Mixture fraction at $m$\\
$\hat{\bm{f}}_{m}$ & Estimated mixture fraction at $m$\\
$\tilde{\bm{f}}_{m}$ & Normalized mixture fraction estimate at $m$\\
${f}_{m,i}$ & Entry of $\bm{f}_{m}$\\
${f}_{\mathrm{th}}$ & Mixture fraction threshold\\
$\bm{w}_{m,j}$ & FO at $m$\\
$j$ & Index for FOs\\
$\mathcal{W}_{m}$ & The set of all FOs at $m$\\
$\bm{u}_{m,p}$ & Likely FO at $m$\\
$p$ & Index for likely FOs\\
$\mathcal{U}_{m}$ & The set of all likely FOs at $m$\\
$\mathbf{C}_{m}$ & Weighting matrix at $m$\\
$C_{m,i}$ & Diagonal entry of $\mathbf{C}_{m}$\\
$\alpha,\beta,\mu$ & Parameters in FORNI\\
$r_{m,n}(i)$ & Basis-neighbor similarity\\
$R_{m}(i)$ & Aggregate basis-neighbor similarity\\
$\theta_{R}$ & An angle threshold\\
$t$ & Index for iterations\\
$N_{\mathrm{p}}$ & Number of voxels processed in parallel\\
\hline
\hline
\end{tabular}
}
\label{tab:symbols}
\end{table}

\subsubsection{FO Estimation with Known Neighborhood Information}
\label{sec:FORNI}

First, we consider a simplified case where the mixture fractions
$\bm{f}_m$  are to be estimated in the voxel $m$ and the mixture
fractions are known at all voxels in a neighborhood $\mathcal{N}_m$ of
voxel $m$. Let $n$ be a 
voxel in $\mathcal{N}_{m}$ and let voxel $n$ have mixture
fractions $\bm{f}_{n}$; then the FOs at voxel $n$ are given by the set 
\begin{equation}
\mathcal{W}_n = \{ \bm{v}_i \, | f_{n,i}
> f_{\mathrm{th}}, i=1,\ldots,N \} \, , 
\label{eq:neighborFOs}
\end{equation}
where $f_{n,i}$ is the $i$-th element of $\bm{f}_n$, and $f_{\mathrm{th}}$ is the threshold for mixture fractions. We further let $\mathcal{W}_{n} = \{\bm{w}_{n,j}\}_{j=1}^{W_{n}}$, where $\bm{w}_{n,j}$ is the $j$-th FO in voxel $n$ and $W_{n}$ is the cardinality of $\mathcal{W}_{n}$. 
For concreteness, we assume at this stage and for the remainder of the
paper that the neighborhood consists of the nearest 26 neighbors~\citep{auria} and $f_{\mathrm{th}} = 0.1$~\citep{landman}.
We want to estimate the mixture fractions $\bm{f}_{m}$ (and therefore the
associated FOs $\mathcal{W}_{m}=\{\bm{w}_{m,j}\}_{j=1}^{W_{m}}$ using Eq.~(\ref{eq:neighborFOs})) at voxel $m$ given the neighborhood information.

Our main goal is to use the patterns of FOs in the neighboring
voxels to encourage a similar pattern of FOs in voxel $m$---this is
the idea of FO smoothness or coherence. 
A set $\mathcal{U}_m$ of \textit{likely FOs} for the voxel $m$ can be computed
given knowledge of the mixture fractions and FOs in its neighbors (the details will be described later).
The likely FO information is then used to influence the mixture fraction
estimation in voxel $m$ using the sparse estimation framework
previously described (for the incorporation of prior knowledge in FO
estimation) in the FIEBR algorithm~\citep{CMIG}. In particular, we solve 
the following weighted $\ell_{1}$-norm regularized least
squares problem
\begin{eqnarray}
\hat{\bm{f}}_{m} = \argmin\limits_{\bm{f}_{m}\geq
  \bm{0}}||\textbf{G}\bm{f}_{m}-\bm{y}_{m}||_{2}^{2} + \beta
||\textbf{C}_{m}\bm{f}_{m}||_{1}\,,
\label{equ:obj2_single}
\end{eqnarray}
where $\textbf{C}_{m}$ is a diagonal matrix that weights the basis
directions according to their distance to the likely FOs in
$\mathcal{U}_m$. For example, the basis directions $\bm{v}_{i}$ that
are closer to the likely FOs in $\mathcal{U}_m$ have smaller weights
in the weighted $\ell_{1}$-norm; therefore, they have smaller penalty in
the objective function and are more likely to be selected as FOs in
$m$. 

\citet{CMIG} specified the diagonal entries of $\textbf{C}_{m}$
as 
\begin{eqnarray}
C_{m,i} = 1 - \alpha\max\limits_{p=1,\ldots,U_m} |\bm{v}_{i}\cdot \bm{u}_{m,p}|,
\quad i=1,\ldots, N \,, 
\label{equ:C_BAMBI}
\end{eqnarray}
where $\bm{u}_{m,p} \in \mathcal{U}_m$, $U_m$ is the cardinality of
$\mathcal{U}_m$, and $\alpha\in[0,1)$ is a constant. Since $\bm{v}_{i}$ and
$\bm{u}_{m,p}$ are unit vectors, $|\bm{v}_{i}\cdot
\bm{u}_{m,p}|\in[0,1]$ and $C_{m,i}$ is positive for all $i$.  
In this work, we find it useful to normalize these diagonal entries so
that the weights on the most likely FOs are nearly the same as when
there is no neighboring FO information used at all. Accordingly, here
we specify the diagonal weights in $\textbf{C}_m$ as 
\begin{equation}
C_{m,i} = \frac{1 - \alpha\max\limits_{p=1,\ldots,U_m} |\bm{v}_{i}\cdot
  \bm{u}_{m,p}|}{\min\limits_{q=1,\ldots,N}\left(1 -
    \alpha\max\limits_{p=1,\ldots,U_m}|\bm{v}_{q}\cdot \bm{u}_{m,p}|\right)}, 
\quad i=1,\ldots, N \,.  
\label{eq:diagonalweights}
\end{equation}
We see that the weights in Eq.~(\ref{eq:diagonalweights}) are just the
weights in Eq.~(\ref{equ:C_BAMBI}) normalized by the smallest
diagonal entry. Note that we require $\alpha\in[0,1)$ in order
to ensure that $C_{m,i}>0$, $\forall i$.   

\citet{CMIG} developed the above framework to incorporate 
fixed prior directions at each voxel in the estimation of FOs. These prior
directions were either hand-drawn or determined by atlas registration.
To apply this framework to spatial smoothness, we replace the concept
of prior directions with that of likely FOs. Note that our application
of FO estimation with spatial coherence is fundamentally different
than~\cite{CMIG} in two respects: 1)~likely FOs are computed based on
neighbors and no manual intervention or anatomical atlas registration
is needed; 2)~because computations of FOs depend on neighbors, the FOs for all voxels need to be simultaneously estimated and the independent FO computation at each voxel in Eq.~(\ref{equ:obj2_single}) is inappropriate. The proposed approach to the computation of likely FOs and simultaneous FO estimation is presented next in Sections~\ref{sec:likely_FO} and \ref{sec:FO_est}, respectively. 

\subsubsection{Computation of Likely FOs from Neighbors}
\label{sec:likely_FO}

\begin{figure}[t]
\centering
   \includegraphics[width=0.98\linewidth]{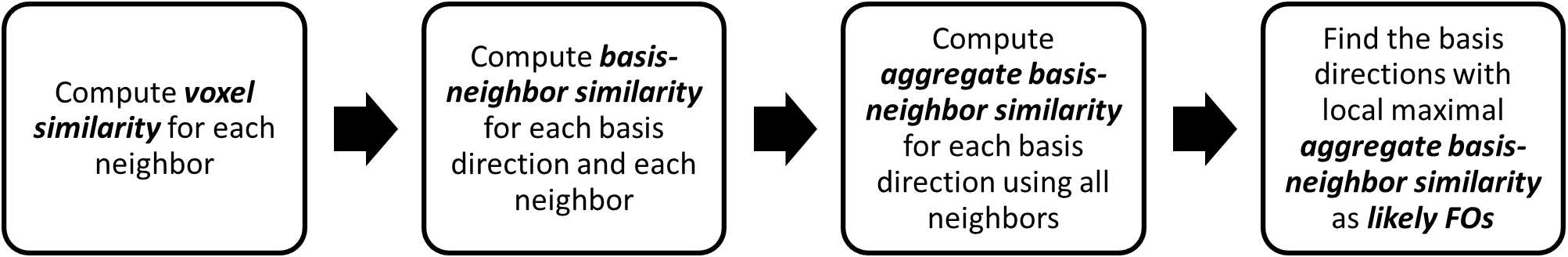}
\caption[]{A flow chart of the likely FO computation at a voxel.} 
\label{fig:flow}
\end{figure}

A flow chart of the likely FO computation at each voxel is shown in Figure~\ref{fig:flow}.
We first consider a single neighbor
voxel; in particular, let voxel $n$ be the neighbor of voxel
$m$. Let $\textbf{D}_{m}$ and $\textbf{D}_{n}$ be the diffusion
tensors fit from diffusion weighted signals at voxels $m$ and $n$,
respectively. Based on the tensors, we define the \textit{voxel similarity} $w_{m,n}$ 
between voxels $m$ and $n$ as
\begin{eqnarray}
w_{m,n} = e^{-\mu d^2(\textbf{D}_{m},\textbf{D}_{n})},
\end{eqnarray}
where $\mu$ is a constant and $d(\cdot,\cdot)$ is the measure of distance between tensors given
by~\citet{arsigny} 
\begin{eqnarray}
d(\textbf{D}_{m},\textbf{D}_{n}) =
\sqrt{\mathrm{Trace}(\{\log(\textbf{D}_{m})-\log(\textbf{D}_{n})\}^{2})}. 
\end{eqnarray}

Given the definition of voxel similarity, we now want a measure of
the similarity of each basis direction $\bm{v}_{i}$ to the FOs $\mathcal{W}_n=\{\bm{w}_{n,j}\}_{j=1}^{W_{n}}$ in voxel $n$. 
Accordingly, we define the \textit{basis-neighbor similarity} $r_{m,n}(i)$ as 
\begin{eqnarray}
r_{m,n}(i) = w_{m,n}\max\limits_{j=1,\ldots,W_n} |\bm{v}_{i}\cdot \bm{w}_{n,j}|, 
\quad i=1,\ldots,N \,.
\label{equ:bn_sim}
\end{eqnarray}
In order for a given basis vector to be similar to a
neighbor's computed FO, the voxels must be similar and the directions
must be well-aligned.

Now consider all voxels that are neighbors of voxel $m$. We define an 
\textit{aggregate basis-neighbor similarity} $R_m(i)$ at voxel $m$ for each
basis vector $\bm{v}_{i}$ as   
\begin{equation}
R_{m}(i) = \sum_{n \in \mathcal{N}_{m}} r_{m,n}(i)\,,  
\quad i=1,\ldots,N \,.
\label{equ:abn_sim}  
\end{equation}
Basis directions with larger aggregate basis-neighbor similarity
correspond to directions that are close to FOs in many neighboring
voxels or in a few neighbors that have strong voxel
similarity. These are directions that are more likely to be FOs in
voxel $m$ by virtue of the current FOs in neighboring voxels.  

\begin{figure}[t]
  \centering
	\subfigure[FO Structure]{
		\includegraphics[width=0.22\columnwidth]{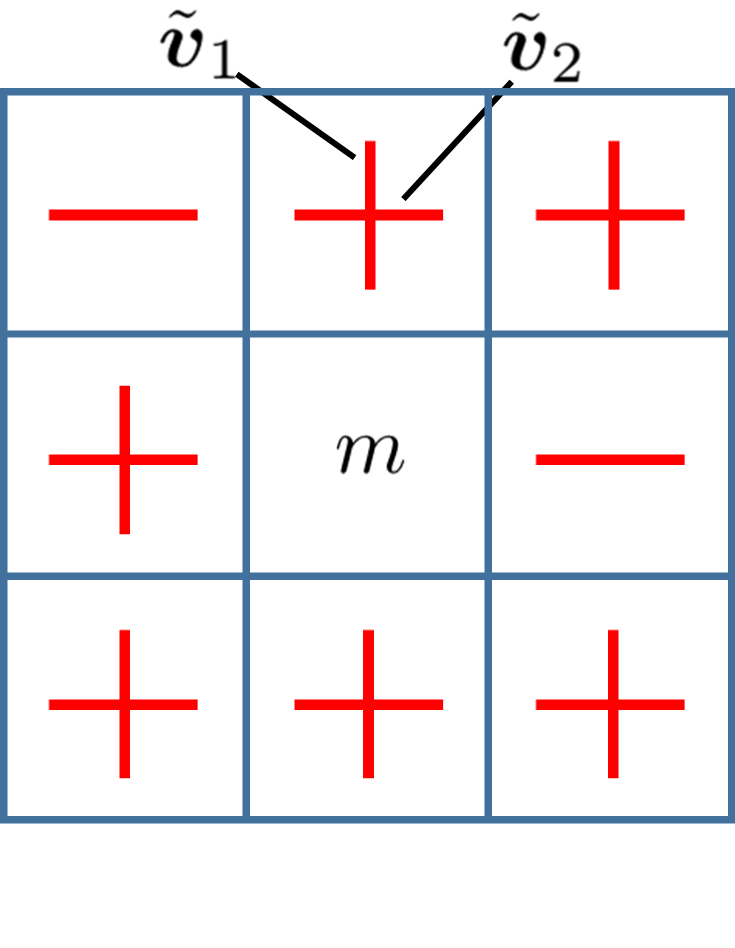}
		\label{fig:sim_illua}
	}
  \subfigure[$R_{m}(i)$ Profile]{
		\includegraphics[width=0.35\columnwidth]{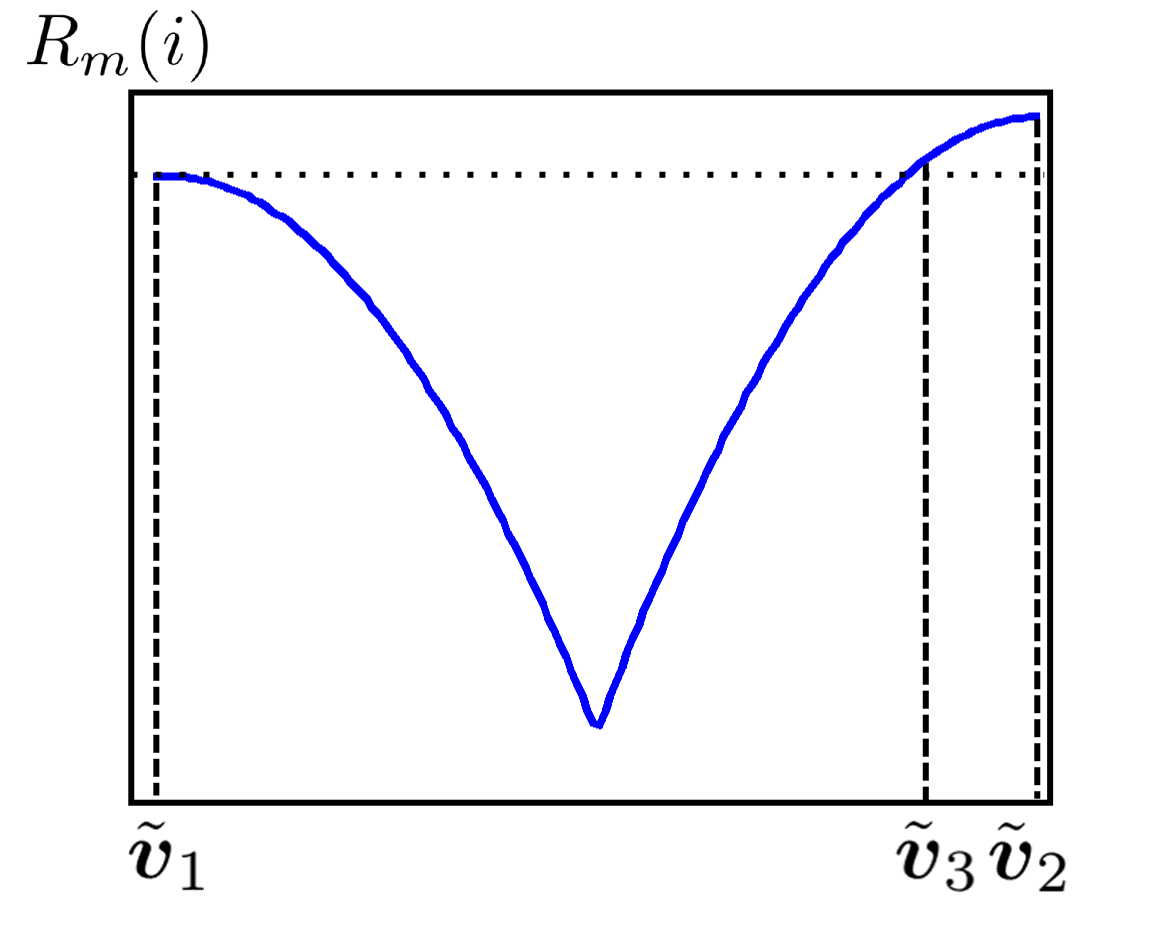}
		\label{fig:sim_illub}
	}
\caption{An illustration of the $R_{m}(i)$ profile in the case where
  one of the crossing FOs fails to be reconstructed in some neighboring
  voxels.} 
\label{fig:sim_illu}
\end{figure}

Given $R_{m}$ we can extract a set of likely FOs for voxel $m$. 
We could simply choose the directions with the largest $R_{m}$
values or those with $R_{m}$ values greater than a threshold. 
However, a special circumstance should be noted. In a crossing region,
such as that depicted in Figure~\ref{fig:sim_illua},
some of the neighboring voxels could fail to estimate a crossing FO due to
noise. In particular, suppose there are two crossing FOs $\tilde{\bm{v}}_{1}$ and 
$\tilde{\bm{v}}_{2}$ in this region, and $\tilde{\bm{v}}_{1}$ fails to be
reconstructed in more than one neighboring voxel while the other
neighbors have both FOs reconstructed. 
In this case, as is shown in Figure~\ref{fig:sim_illub}, the basis vector $\bm{v}_{i}$ corresponding to
$\tilde{\bm{v}}_{1}$ 
could have a smaller $R_{m}(i)$ value than a basis direction
$\tilde{\bm{v}}_{3}$, for example, which is distant from both
$\tilde{\bm{v}}_{1}$ and $\tilde{\bm{v}}_{2}$. In this case, the desired $\tilde{\bm{v}}_{1}$ is not included in the top two likely FOs and thus $\tilde{\bm{v}}_{1}$ may not be properly encouraged, while an undesired $\tilde{\bm{v}}_{3}$ may be encouraged and create false FOs.

A more robust definition of the likely FOs is as those directions
which comprise the local maxima of $R_m$.  In particular, consider
the direction $\bm{v}_i$ and compare its value $R_m(i)$ to
all values of $R_m(i')$ corresponding to directions $\bm{v}_{i'}$ within $\theta_{R} =
20^{\circ}$. If $R_m(i)$ is maximum, then $\bm{v}_i$ is included in
the likely FOs. Following this reasoning, the likely FOs at voxel $m$ are given by  
\begin{eqnarray}
\mathcal{U}_m =\{\bm{v}_{i}| \forall\:i' \neq i \mbox{ and} \arccos(|\bm{v}_{i}\cdot\bm{v}_{i'}|)\leq
\frac{\pi}{180^\circ}\theta_{R}: R_{m}(i)\geq R_{m}(i')\}.
\label{equ:likely_FOs} 
\end{eqnarray}
Note that $\theta_{R}$ is converted to have a unit of radians so that it can be compared with $\arccos(|\bm{v}_{i}\cdot\bm{v}_{i'}|)$.

\begin{figure}[t]
\centering
   \includegraphics[width=0.65\linewidth]{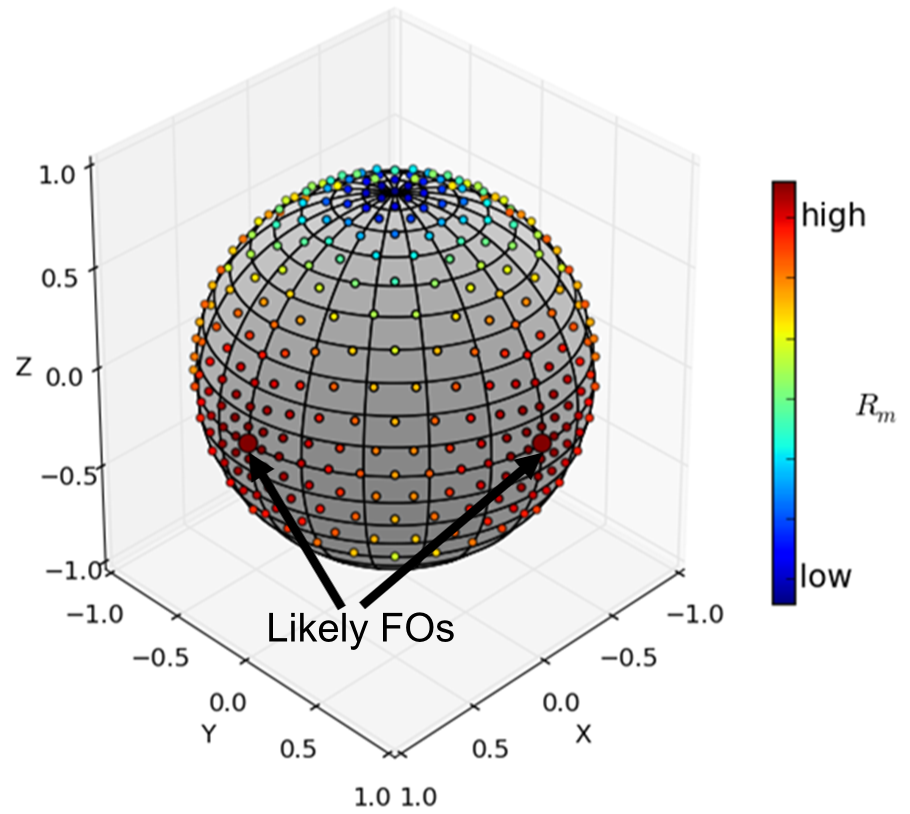}
\caption[]{An example of the $R_{m}$ values of a voxel in the crossing region in Figure~\ref{fig:toy}. Each $R_{m}(i)$ is plotted on the unit sphere according to its associated basis direction $\bm{v}_{i}$. The $R_{m}$ values of the likely FOs are indicated by the larger dots and black arrows.} 
\label{fig:Rm}
\end{figure}

An example of the $R_{m}$ values of a voxel in the crossing region in Figure~\ref{fig:toy} is shown in Figure~\ref{fig:Rm}, where the $R_{m}$ values are plotted on the unit sphere according to their associated basis directions. The $R_{m}$ values of the likely FOs are indicated by the larger dots and black arrows. The two likely FOs are the horizontal ($x$) and vertical~($y$) directions in Figure~\ref{fig:toy} that correspond to the desired FOs.

\subsubsection{FO Estimation for All Voxels}
\label{sec:FO_est}

After the likely FOs for voxel $m$ are determined, the weighting matrix
$\textbf{C}_{m}$ can be obtained.  Note that in
Eq.~(\ref{equ:obj2_single}), we have assumed known neighbor
information. However, the FOs in the neighboring voxels are also unknown
and must be estimated, which means that FOs cannot be estimated
independently---it is a joint FO estimation problem. 

Suppose the total number of voxels of interest is $M$. The estimation
of all unknown mixture fractions
$\bm{f}=(\bm{f}_{1}^{T},\bm{f}_{2}^{T},\ldots,\bm{f}_{M}^{T})^{T}$
(through which the FOs are also estimated) can be written as 
\begin{eqnarray}
\hat{\bm{f}} = \argmin\limits_{\bm{f} \geq \bm{0}} E(\bm{f}) =
\argmin\limits_{\bm{f}_{1},\bm{f}_{2},\ldots,\bm{f}_{M}\geq
  \bm{0}}\sum\limits_{m=1}^{M}||\textbf{G}\bm{f}_{m}-\bm{y}_{m}||_{2}^{2}
+ \beta ||\textbf{C}_{m}\bm{f}_{m}||_{1} \,.  
\label{equ:obj2_all}
\end{eqnarray}
Here $\textbf{C}_{m}$ encodes the interaction between
neighbors. 
Since $\textbf{C}_{m}$ depends on the parameter $\alpha$ (see  
Eq.~(\ref{eq:diagonalweights})), $\alpha$ and $\beta$ are the 
two parameters that must be specified by the user. When $\alpha$
is larger there is more influence from neighboring 
voxels, and when $\beta$ is larger the mixture fractions tend to be
more sparse and therefore there are fewer estimated FOs. 

\subsection{Minimization of the Objective Function and Parallelization}

\algblock{ParFor}{EndParFor}
\algnewcommand\algorithmicparfor{\textbf{parfor}}
\algnewcommand\algorithmicpardo{\textbf{do}}
\algnewcommand\algorithmicendparfor{\textbf{end\ parfor}}
\algrenewtext{ParFor}[1]{\algorithmicparfor\ #1\ \algorithmicpardo}
\algrenewtext{EndParFor}{\algorithmicendparfor}

\begin{algorithm}
\caption{FORNI}
	\label{alg:para}
	\begin{spacing}{1.0}
	\begin{algorithmic}[1]
  \Require  {Diffusion weighted signals
    $\{S_{m}(\bm{q}_{1}),\ldots,S_{m}(\bm{q}_{K})\}_{m=1}^{M}$;
    baseline signals $\{S_{0m}\}_{m=1}^{M}$ without diffusion
    weighting; diffusion gradients $\{\bm{q}_{1},\ldots,\bm{q}_{K}\}$;
    size of parallel processing: $N_{\mathrm{p}}$; the tensor basis
    $\{\textbf{D}_{i}\}_{i=1}^{N}$ and their PEVs
    $\{\bm{v}_{i}\}_{i=1}^{N}$; maximum number of iterations
    $t_{\mathrm{max}}$; the initialization of FOs $\{\mathcal{W}_{m}^{0}\}_{m=1}^{M}$ computed
    from~\citet{landman}; the iteration number starts from $t=1$;} 
	\Ensure  {Mixture fractions $\{\bm{f}_{m}\}_{m=1}^{M}$ and FOs $\{\mathcal{W}_{m}\}_{m=1}^{M}$}
	\State{Compute the attenuation matrix $\textbf{G}$: $G_{ki}=e^{-\bm{q}_{k}^{T}\textbf{D}_{i}\bm{q}_{k}}$
	}
	\State{Compute $\{\bm{y}_{m}\}_{m=1}^{M}$: $\bm{y}_{m}=(S_{m}(\bm{q}_{1})/S_{0m},\ldots,S_{m}(\bm{q}_{K})/S_{0m})^{T}$}
	\State{Initialize FOs: $\{\mathcal{W}_{m}\}_{m=1}^{M} \coloneqq \{\mathcal{W}_{m}^{0}\}_{m=1}^{M}$}
	\While{$t \leq t_{\mathrm{max}}$}
	\For{$a=0:\left\lceil\frac{M}{N_{\mathrm{p}}}-1\right\rceil$} 
	\ParFor{$b=1:\min({N_{\mathrm{p}},M-aN_{\mathrm{p}}})$}
	\State{$m \coloneqq aN_{\mathrm{p}}+b$}
	\ForAll{basis directions $\bm{v}_{i}$} 
	\State \parbox[t]{\dimexpr\linewidth-\algorithmicindent}{
	$R_{m}(i) \coloneqq \sum\limits_{n\in
          \mathcal{N}_{m}}w_{m,n}\max\limits_{j=1,\ldots,W_{n}} |\bm{v}_{i}\cdot \bm{w}_{n,j}|$} 
	\EndFor
	\State \parbox[t]{\dimexpr\linewidth-\algorithmicindent}{
	$\{\bm{u}_{m,p}\}_{p=1}^{U_{m}} \coloneqq \{\bm{v}_{i}| \forall\:i' \neq i \mbox{ and} \arccos(|\bm{v}_{i}\cdot\bm{v}_{i'}|)\leq \frac{\pi}{180^\circ}\theta_{R}: R_{m}(i)\geq R_{m}(i') \}$}
	\For{$i=1:N$} 
	\State \parbox[t]{\dimexpr\linewidth-\algorithmicindent}{$C_{m,i} \coloneqq {\scriptstyle \left(1 - \alpha\max\limits_{p=1,\ldots,U_{m}}|\bm{v}_{i}\cdot \bm{u}_{m,p}|\right)\Big/\left(\min\limits_{q=1,\ldots,N}\left(1 - \alpha\max\limits_{p=1,\ldots,U_{m}}|\bm{v}_{q}\cdot \bm{u}_{m,p}|\right)\right)}$}
	\EndFor
	\State{Solve Eq.~(\ref{equ:bcd2}) to obtain $\hat{\bm{f}}_{m}^{t}$}
	\State{$\bm{f}_{m} \coloneqq \hat{\bm{f}}_{m}^{t}/||\hat{\bm{f}}_{m}^{t}||_{1}$}
	\EndParFor
	\For{$b=1:\min({N_{\mathrm{p}},M-aN_{\mathrm{p}}})$}
	\State{$m \coloneqq aN_{\mathrm{p}}+b$}
	\State{$\mathcal{W}_{m} \coloneqq \{\bm{v}_{i}|f_{m,i} > f_{\mathrm{th}}, i=1,\ldots,N\}$}
	\EndFor
	\EndFor
	\State{$t \coloneqq t+1$ \textbf{until convergence}}
	\EndWhile
	\State{\textbf{return} $\{\bm{f}_{m}\}_{m=1}^{M}$ and $\{\mathcal{W}_{m}\}_{m=1}^{M}$}
	\end{algorithmic}
	\end{spacing}
\end{algorithm}

In Eq.~(\ref{equ:obj2_all}), the FOs in each voxel are coupled with
neighbor voxels in the weighting matrix $\textbf{C}_{m}$. We 
use an iterative \textit{block coordinate descent} (BCD)
method~\citep{bertsekas} to decouple the interaction and optimize the
objective function. At iteration~$t$, each $\bm{f}_{m}$ is estimated
by solving 
\begin{eqnarray}
\hat{\bm{f}}^{t}_{m} &=& \argmin\limits_{\bm{f}_{m} \geq \bm{0}} 
E(\hat{\bm{f}}^{t}_{1},...,\hat{\bm{f}}^{t}_{m-1},\bm{f}_{m},
\hat{\bm{f}}^{t-1}_{m+1},...,\hat{\bm{f}}^{t-1}_{M})   
\\ 
&=& \argmin\limits_{\bm{f}_{m}\geq
  \bm{0}}||\textbf{G}\bm{f}_{m}-\bm{y}_{m}||_{2}^{2} + \beta
||\textbf{C}_{m}^{t}\bm{f}_{m}||_{1}, 
\label{equ:bcd2}
\end{eqnarray}
where $\textbf{C}_{m}^{t}$ is the diagonal weighting matrix at iteration $t$ and it is determined by the likely FOs computed from the neighbor FOs at iteration $t$ or $t-1$ according to Eq.~(\ref{eq:diagonalweights}).
The detailed update of $\textbf{C}_{m}^{t}$ at each iteration and the optimization of Eq.~(\ref{equ:bcd2}) are given in~\ref{sec:appendix}.
Finally, $\hat{\bm{f}}_{m}^{t}$ is normalized so
that entries sum to unity and the FOs $\mathcal{W}_{m}^{t}$ at
voxel $m$ at time $t$ are determined using Eq.~(\ref{eq:neighborFOs}).

Because the $\ell_{1}$-norm regularized least squares problem in
Eq.~(\ref{equ:bcd2}) must be solved for every voxel in each
iteration, the algorithm requires heavy computation and can be
time-consuming. Therefore, we modified the BCD optimization in
Eq.~(\ref{equ:bcd2}) so that multiple voxels can be simultaneously
solved to speed up processing. We process $N_{\mathrm{p}}\geq 2$
voxels together (in this work $N_{\mathrm{p}}=8$). Each voxel $m$ can
be represented as $m=aN_{\mathrm{p}}+b$, where $a$ and $b$ are
integers ($0\leq a < \frac{M}{N_{\mathrm{p}}}$ and $1\leq b \leq
N_{\mathrm{p}}$). 
For each group $aN_{\mathrm{p}}+1 \leq m \leq (a+1)N_{\mathrm{p}}$ with fixed $a$, we have
\begin{eqnarray}
\hat{\bm{f}}^{t}_{m}=
\argmin\limits_{\bm{f}_{m} \geq \bm{0}}
E\left(\{\hat{\bm{f}}^{t}_{m_{0}}\}_{m_{0}\leq
    aN_{\mathrm{p}}},\bm{f}_{m},\{\hat{\bm{f}}^{t-1}_{m_{1}}\}_{m_{1}>aN_{\mathrm{p}},m_{1}\neq
    m}\right) \,,
\label{equ:parallel}
\end{eqnarray}
and these $\bm{f}_{m}$'s can be solved in parallel.

The above iterative algorithm is initialized using
CFARI~\citep{landman}, which provides the mixture fractions
$\hat{\bm{f}}_{m}^{0}$ at each voxel independently. 
The algorithm terminates when the FO difference between two
successive iterations is small or when the maximum number of
iterations is reached. The complete algorithm is summarized in Algorithm~\ref{alg:para}.

\section{Experiments}
\label{sec:exp}

FORNI was evaluated first on a digital crossing phantom, then on
\textit{ex vivo} tongue dMRI data from one subject, and finally on an
\textit{in vivo} brain dMRI dataset comprising six subjects. 
FORNI was compared to the SHORE algorithm~\citep{merlet2013} which estimates the ensemble average propagator and orientation distribution function using the SHORE basis~\citep{cheng2011}, the constrained spherical deconvolution (CSD) algorithm~\citep{tournier2} which introduces a nonnegative constraint on the spherical harmonics framework, the CFARI algorithm that estimates the FOs using a voxelwise sparse reconstruction~\citep{landman}, an FO smoothing algorithm (CFARI-s)~\citep{sigurdsson} that smooths the CFARI results, and an FO estimation algorithm (L2L0NW)~\citep{auria} that uses structured sparsity to enforce smooth FO estimation. In the experiments on the tongue dMRI data, we have also compared our method with the FIEBR algorithm~\citep{CMIG} that is designed for the tongue to improve FO estimation by using atlas information. In the experiments on brain dMRI, because multiple $b$-values were used to acquire the brain dMRI data, the CSD algorithm was
replaced by generalized $q$-sampling imaging (GQI)~\citep{gqi} which
can reconstruct FOs using multi-shell dMRI. SHORE and CSD are implemented using the Dipy software~\citep{DIPY} (\url{http://nipy.org/dipy/documentation.html}); CFARI and CFARI-s are implemented in the JIST software framework~\citep{lucas}; GQI is implemented in the DSI Studio software~(\url{http://dsistudio.labsolver.org/Manual/Reconstruction}); and L2L0NW was performed using the code provided by its authors at \url{https://github.com/basp-group/co-dmri}.

\subsection{3D Digital Crossing Phantom}

\begin{figure}[t]
\centering
   \includegraphics[width=0.7\linewidth]{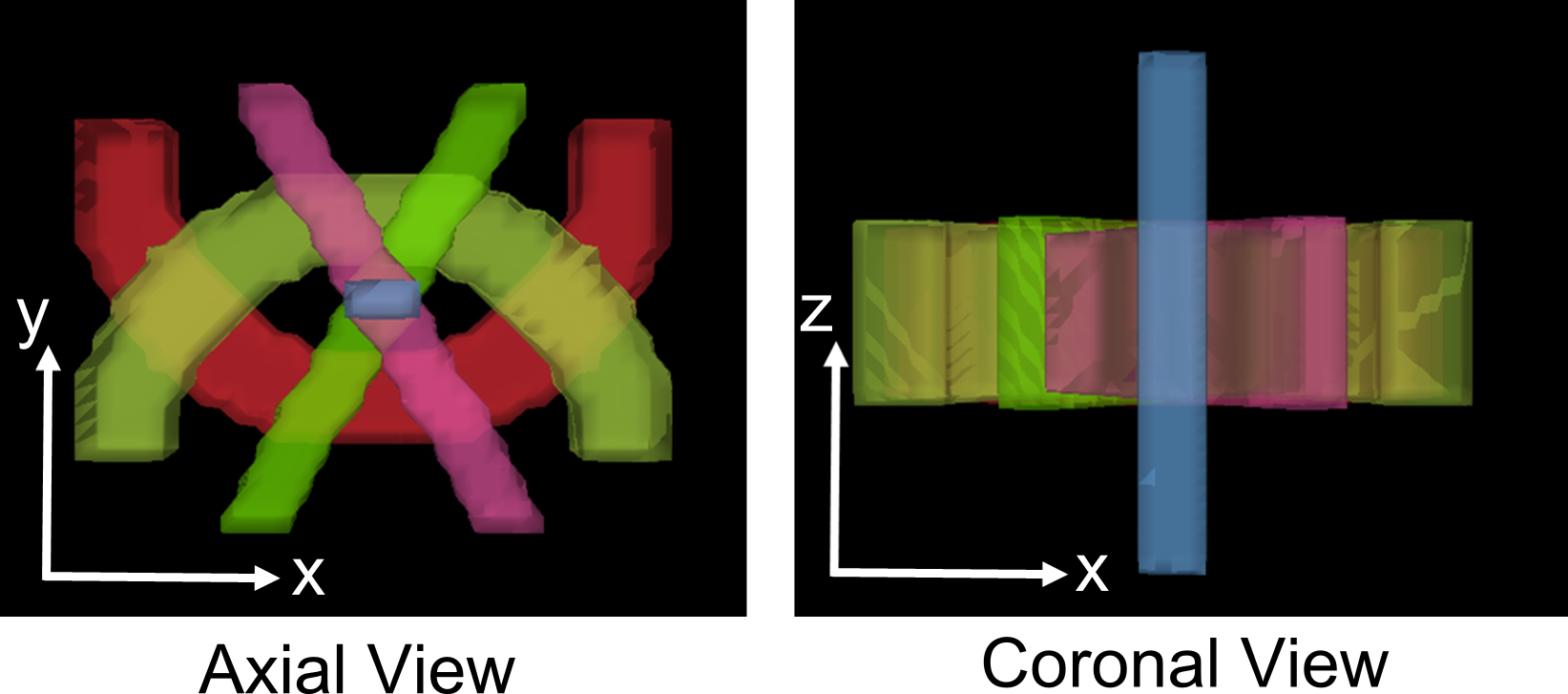}
\caption[]{A 3D rendering of the simulated tracts.} 
\label{fig:phantom}
\end{figure}

\begin{figure}[!t]
\centering
   \includegraphics[width=0.95\linewidth]{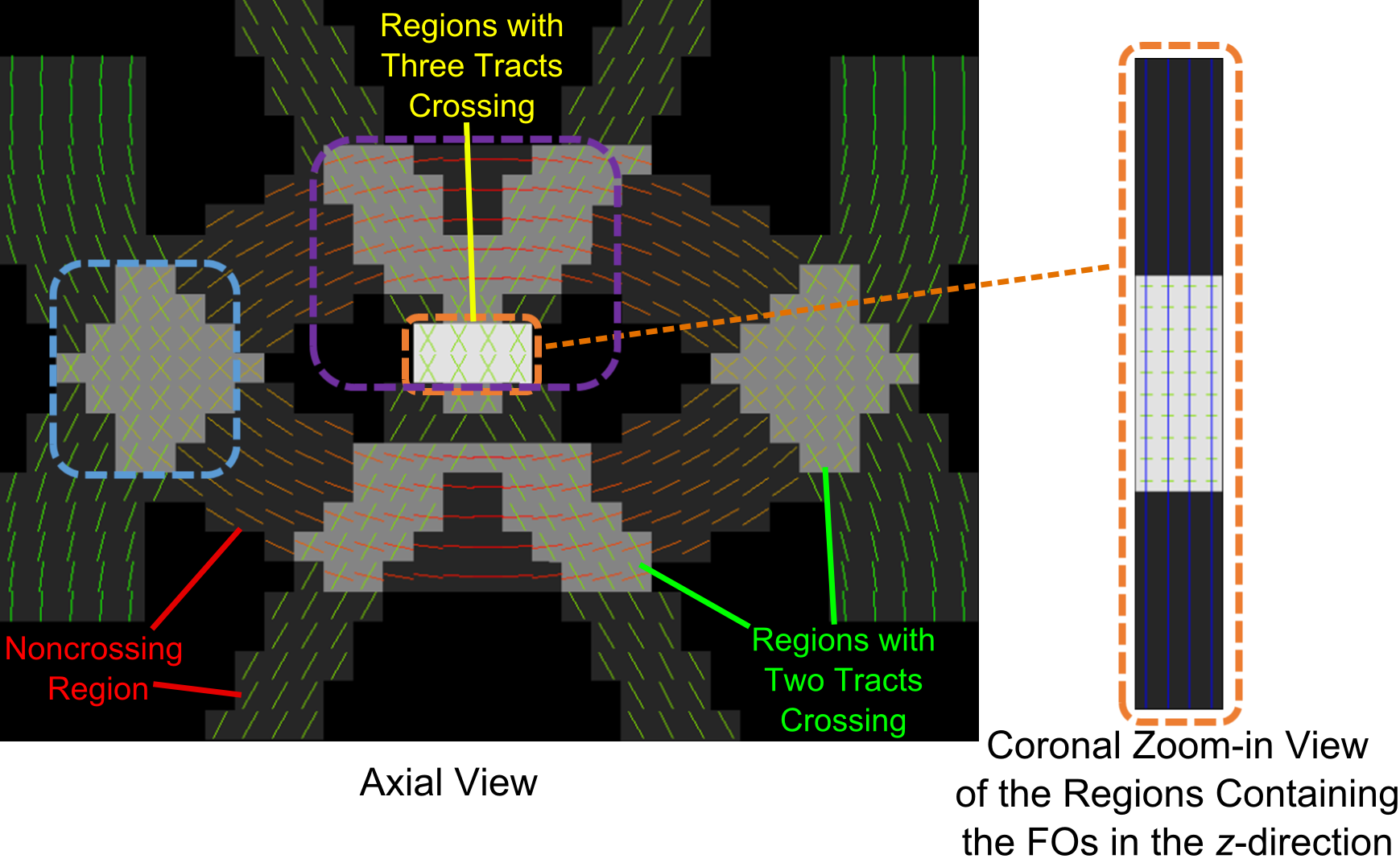}
\caption[]{Ground truth FOs overlaid on the map indicating the number of FOs at each voxel. The FOs in the $z$-direction are shown in a coronal view. The regions highlighted by the dashed boxes are later zoomed in for qualitative evaluation in Figure~\ref{fig:phantom_fo_all}. The visualization of FOs was created in FSLView~\citep{FSL}.} 
\label{fig:phantom_fo}
\end{figure}

A 3D digital crossing phantom (available at \url{https://www.iacl.ece.jhu.edu/Chuyang}) was generated to simulate five fiber tracts (see Figure~\ref{fig:phantom}), where one $b0$ image and 60 gradient directions ($b=1000\,\mathrm{s}/\mathrm{mm}^{2}$) were used. A two-tensor/three-tensor model was used to create the simulated diffusion signals for regions with two/three crossing tracts. The eigenvalues of each individual tensor are $\lambda_{1}=2.0\times 10^{-3}~\mathrm{mm}^{2}/\mathrm{s}$ and $\lambda_{2}=\lambda_{3}=0.5\times 10^{-3}~\mathrm{mm}^{2}/\mathrm{s}$. Thus, for each individual tensor the \textit{fractional anisotropy} (FA) is 0.71 and the \textit{mean diffusivity} (MD) is $1.0\times 10^{-3}~\mathrm{mm}^{2}/\mathrm{s}$.
Rician noise with different \textit{signal-to-noise ratio} (SNR) ($\mathrm{SNR}=10$, $20$, and $30$) on the $b0$ image was added to the \textit{diffusion weighted images} (DWIs).

\begin{figure}[!t]
\centering
   \includegraphics[width=0.87\linewidth]{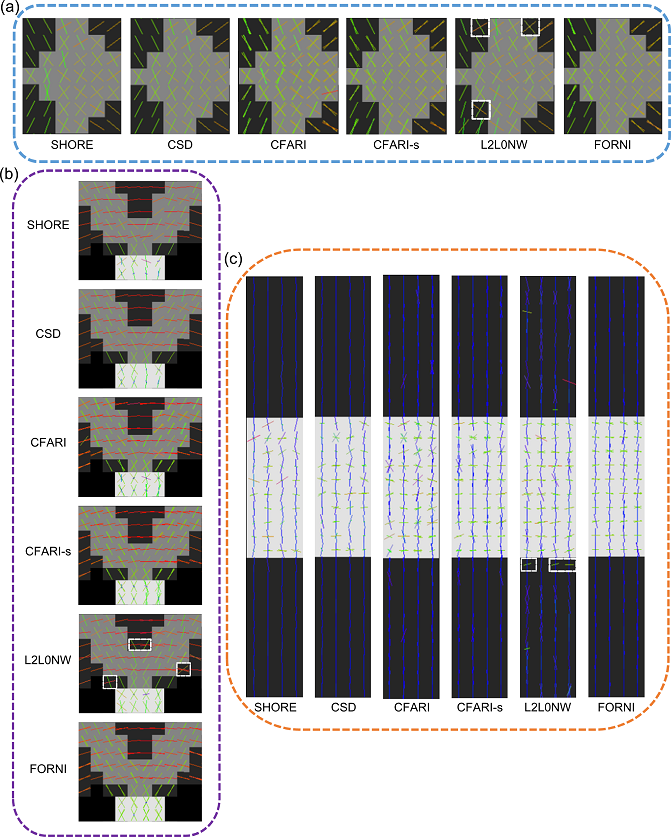}
\caption[]{FO estimation results overlaid on the map indicating the number of ground truth FOs at each voxel at $\mathrm{SNR}=20$ in the regions highlighted in Figure~\ref{fig:phantom_fo}. Note the white boxes where L2L0NW produces false positive FOs. The visualization of FOs was created in FSLView~\citep{FSL}.}
\label{fig:phantom_fo_all}
\end{figure}

FORNI (with $\alpha=0.8$, $\beta=0.5$, and $\mu=3.0$) was applied and compared
with SHORE~\citep{merlet2013}, CSD~\citep{tournier2}, CFARI~\citep{landman}, CFARI-s~\citep{sigurdsson}, L2L0NW~\citep{auria}, and the ground truth. 
The ground truth FOs are shown in Figure~\ref{fig:phantom_fo}. Because the FOs in the $z$-direction are not visible in the axial view, the regions containing these FOs are also shown in the coronal view.
A qualitative evaluation of FO estimation is shown in Figure~\ref{fig:phantom_fo_all} using the results at $\mathrm{SNR}=20$ in the highlighted regions in Figure~\ref{fig:phantom_fo}. The FOs are color-coded by the standard DTI color scheme (red: left--right; green: front--back; and blue:
up--down)~\citep{pajevic} and overlaid on the map indicating the number of ground truth FOs at each voxel. 
We can see that FORNI produces smooth FOs compared with SHORE, CSD, and CFARI which perform voxelwise FO
estimation. 
In the regions (highlighted by the orange box in Figures~\ref{fig:phantom_fo} and \ref{fig:phantom_fo_all}) containing three crossing tracts, both L2L0NW and FORNI are able to better recover the crossing patterns than SHORE, CSD, CFARI, and CFARI-s.
In addition, FORNI does not produce false positive FOs (see the white boxes in Figure~\ref{fig:phantom_fo_all}) as in the results of L2L0NW when neighbor information is used.

To quantitatively evaluate the results, we define a voxelwise error measure of FOs in degrees:
\begin{equation}
e_{\mathrm{FO}} = \max\left(\frac{1}{N_{1}}\sum_{i=1}^{N_{1}}
\min\limits_{j} 
\arccos (|\bm{w}_{i}\cdot \bm{u}_{j}|),
\frac{1}{N_2} \sum\limits_{j=1}^{N_{2}} 
\min\limits_{i} 
\arccos (|\bm{w}_{i}\cdot \bm{u}_{j}|)\right) 
\cdot \frac{180^{\circ}}{\pi}.
\label{eq:eFOdef}
\end{equation}
Here, $\bm{w}_{i}$ and $\bm{u}_{j}$ are the estimated and ground truth
FOs, respectively, and $N_{1}$ and $N_{2}$ are the numbers of $\bm{w}_{i}$
and $\bm{u}_{j}$, respectively.
Note that $\arccos(\cdot)$ is in radians and it is converted to degrees by multiplying $\frac{180^{\circ}}{\pi}$.
In the $\max$ function of Eq.~(\ref{eq:eFOdef}), the first term measures
how far away the estimated FOs are from the true FOs, and the second
term measures how accurate the true FOs are estimated.  
Since both terms are expected to be small when the FO estimation is
accurate, the worst of the two errors is reported. 

\begin{figure}[t]
\centering
		\includegraphics[width=0.65\columnwidth]{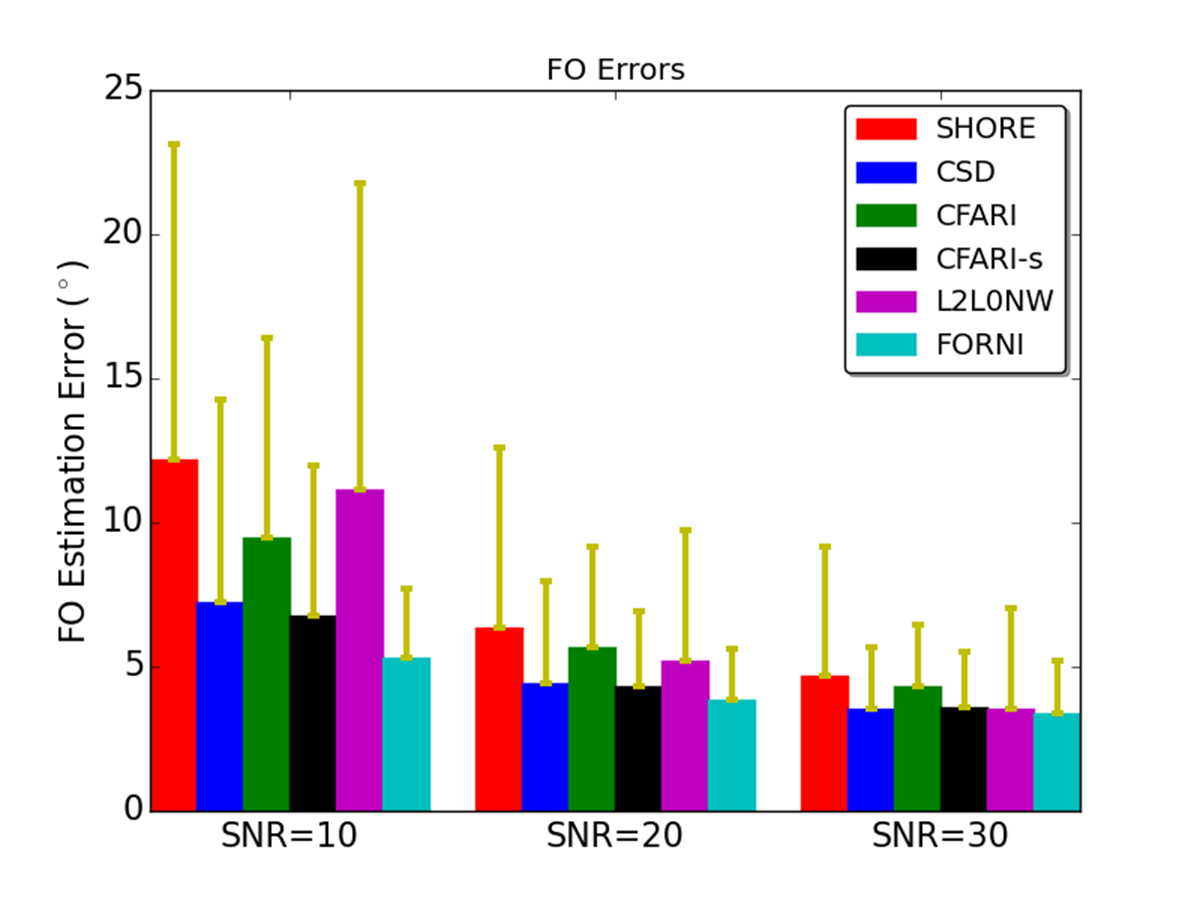}
\caption{
Means and standard deviations of FO errors over the entire phantom at $\mathrm{SNR}=10$, $20$, and $30$.} 
\label{fig:phantom_error}
\end{figure}

We compared the FO errors of FORNI over the entire phantom with those of SHORE, CSD, CFARI, CFARI-s, and L2L0NW using all three noise levels. The results are shown in Figure~\ref{fig:phantom_error}, where means and standard deviations of the FO errors are plotted. It can be seen for all methods the FO errors increase as SNR decreases. FORNI produces more accurate FOs than the competing methods in all three cases.

\begin{figure}[!t]
  \centering
		\includegraphics[width=0.9\columnwidth]{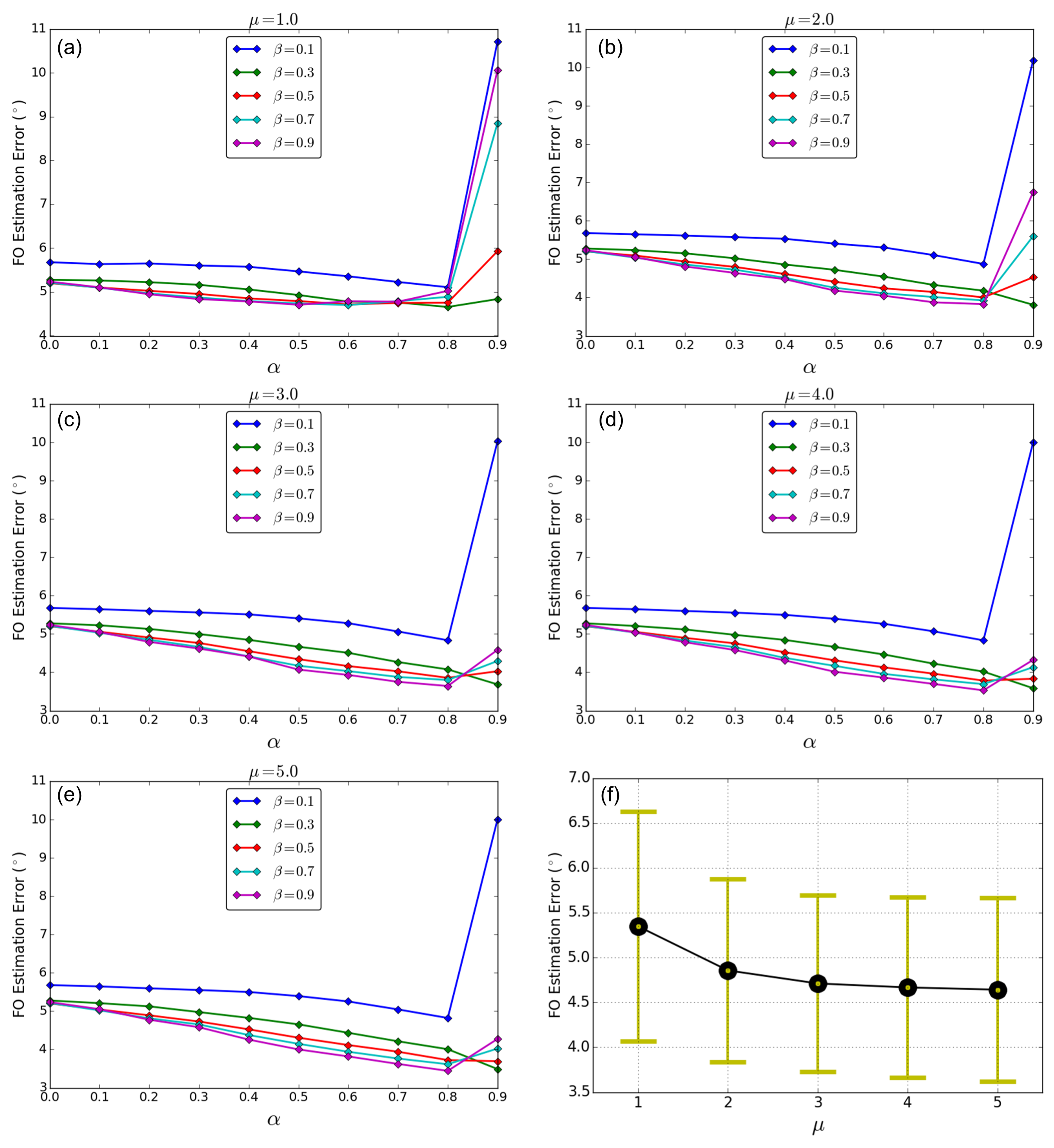}
\caption{Average FORNI FO errors over the entire phantom at $\mathrm{SNR}=20$ with different $\alpha$, $\beta$, and $\mu$: (a)--(e) $\mu=1.0,2.0,\ldots,5.0$; (f) means and standard deviations of the data points in (a)--(e) at different $\mu$.} 
\label{fig:phantom_para}
\end{figure}

\begin{figure}[!t]
  \centering
		\includegraphics[width=0.9\columnwidth]{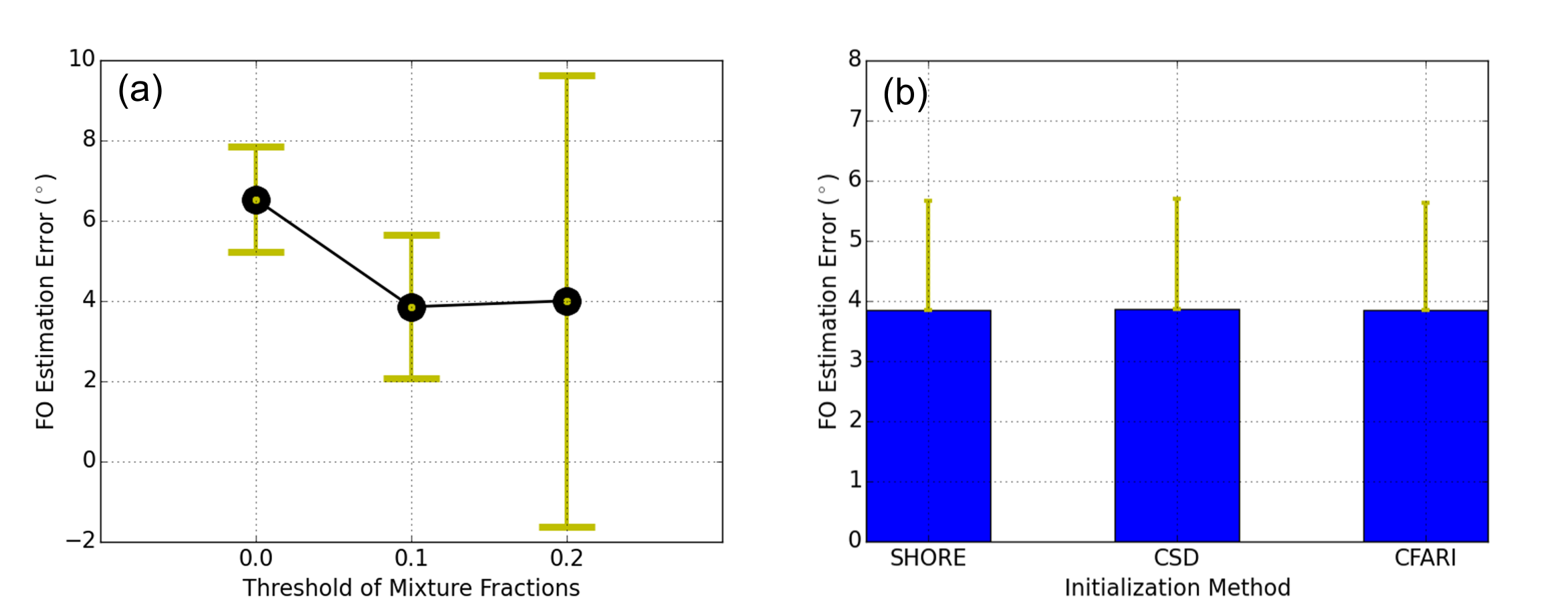}
\caption{Means and standard deviations of FORNI FO errors at $\mathrm{SNR}=20$ with different (a) mixture fraction thresholds and (b) initialization methods.} 
\label{fig:phantom_error_init_f}
\end{figure}

Next, we studied the effect of the parameters, mixture fraction thresholds, and initialization in FORNI. Because in the tongue and brain dMRI data used in this work, the SNR is close to $20$ on the $b0$ images, we used the phantom at $\mathrm{SNR}=20$ for the evaluation below.

To evaluate the impact of parameters, we experimented with different $\alpha$, $\beta$, and $\mu$ settings: $\alpha\in \{0.0,0.1,\ldots,0.9\}$, $\beta\in \{0.1,0.3,\ldots,0.9\}$, and $\mu \in \{1.0,2.0,\ldots,5.0\}$. The average FO errors for each $\alpha$, $\beta$, and $\mu$ combination over the entire phantom are plotted in Figures~\ref{fig:phantom_para}(a)--(e). 
Note that when $\alpha=0.0$, the basis directions are uniformly weighted and no neighbor information is used, which is equivalent to the CFARI algorithm.
At each $\mu$ for most $\beta$, increasing $\alpha$ reduces FO errors until $\alpha$ is too large, and in most cases incorporation of neighborhood information ($\alpha>0$) improves the estimation accuracy. 
Figure~\ref{fig:phantom_para}(f) gives the means and standard deviations of these average FO errors using all $\alpha$ and $\beta$ combinations at each $\mu$. 
It can be seen that the error starts to become stable when $\mu$ reaches 3.0. In addition, the variance is smallest at $\mu=3.0$, indicating the performance is less affected by changing $\alpha$ and $\beta$ values than at other $\mu$. Therefore, for our real data application, we use $\mu=3.0$. Then, at $\mu=3.0$ we select $\alpha$ and $\beta$ for the real data experiments with the following rationale.
First, because in many cases $\alpha=0.9$ causes errors even larger than those without neighbor information, we only consider the choices with $\alpha<0.9$. Second, we have found from experience that choosing $\beta$ too large can lead to instabilities in the $\ell_1$ solver. Therefore, we have picked $\alpha = 0.8$ and $\beta = 0.5$ for the remaining experiments. From Figure~\ref{fig:phantom_para}(c), we see that at $\mu=3.0$ this selection yields a
performance that is comparable to the other four top performing operating points. 

Using the selected $(\alpha,\beta,\mu)=(0.8,0.5,3.0)$, we computed the FO errors with different mixture fraction thresholds ($f_\mathrm{th}\in\{0.0,0.1,0.2\}$) and different initialization methods including SHORE, CSD, and CFARI, which are the voxelwise FO estimation algorithms evaluated in this work. The results are shown in Figure~\ref{fig:phantom_error_init_f}. In Figure~\ref{fig:phantom_error_init_f}(a), we can see that $f_\mathrm{th}=0.1$ (as selected in Section~\ref{sec:FORNI}) achieves the smallest error. In Figure~\ref{fig:phantom_error_init_f}(b), different initialization methods achieve very close FO errors in FORNI, which indicates the robustness of FORNI to initialization.

\subsection{\textit{Ex Vivo} Tongue dMRI}

\begin{figure}[t]
  \centering
  \includegraphics[width=0.9\columnwidth]{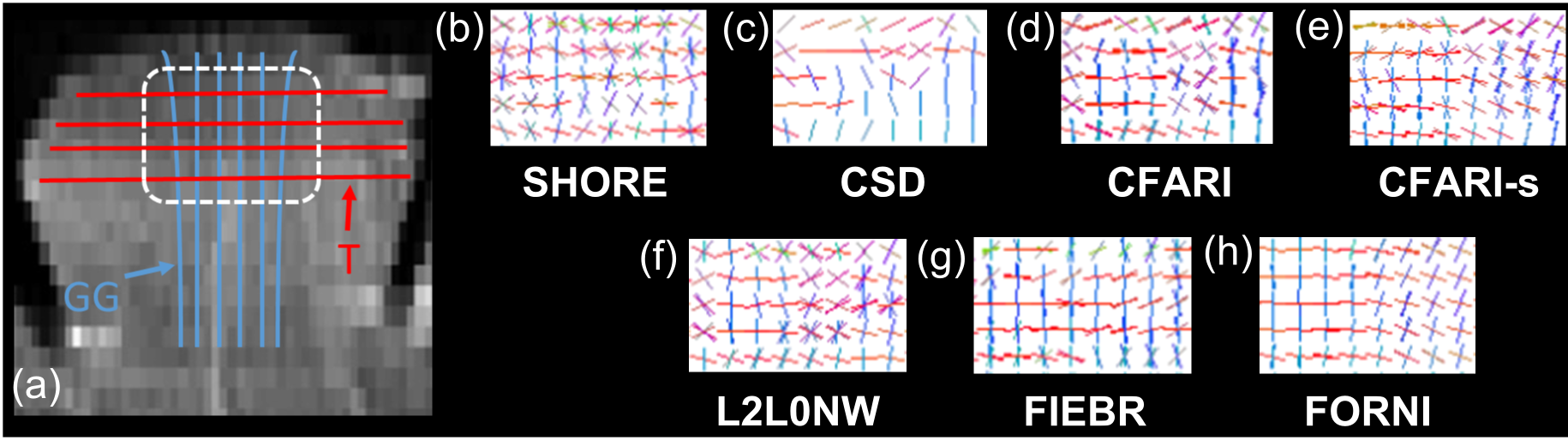}
\caption{FO estimation on the \textit{ex vivo} tongue in the coronal
  view, which is focused on the crossing (highlighted by the white dashed box in (a)) of the GG and T muscle. A high resolution structural image (left) with a schematic of the anatomy of GG and T is shown for location reference.}
\label{fig:tongue}
\end{figure}

Next, FORNI (with $\alpha=0.8$, $\beta=0.5$ and $\mu=3.0$) was applied to the
\textit{ex vivo} tongue dMRI data. Nine $b0$ images and 64 DWIs
($b=2000~\mathrm{s}/\mathrm{mm}^{2}$) were acquired on a 3T MRI
scanner (Magnetom Trio, Siemens, Erlangen, Germany). The resolution is
2~mm isotropic (matrix size: $100\times 100$) and the number of slices
is 30. 
The SNR on the $b0$ image is approximately $20.31$, which was estimated by placing bounding boxes in the background and tract regions~\citep{CMIG}.
Eddy current correction and diffusion tensor estimation were
performed by CATNAP~\citep{CATNAP} implemented in the JIST
software~\citep{lucas}. According to the tensors in noncrossing regions, the eigenvalues of the basis tensors
used in FORNI were $\lambda_{1}=7.0\times
10^{-4}~\mathrm{mm}^{2}/\mathrm{s}$ and $\lambda_{2}=\lambda_{3}=3.0
\times 10^{-4}~\mathrm{mm}^{2}/\mathrm{s}$. The FORNI processing took
around seven minutes for the data. 
SHORE~\citep{merlet2013}, CSD~\citep{tournier2},
CFARI~\citep{landman}, \mbox{CFARI-s}~\citep{sigurdsson}, L2L0NW~\citep{auria}, and FIEBR~\citep{CMIG} were also applied on the data.

Figure~\ref{fig:tongue} shows results in the coronal view with the
focus on the crossing region of the genioglossus (GG) and
the transverse (T) muscle in the tongue. 
A high resolution structural image with a schematic of the anatomy of GG and T is shown in Figure~\ref{fig:tongue}(a) for location reference.
It can be seen that FORNI not only produces smoother FOs but also
better reconstructs the crossing FOs of GG and T. 

\subsection{\textit{In Vivo} Brain dMRI}

\begin{figure}[!t]
		\centering
		\includegraphics[width=0.9\columnwidth]{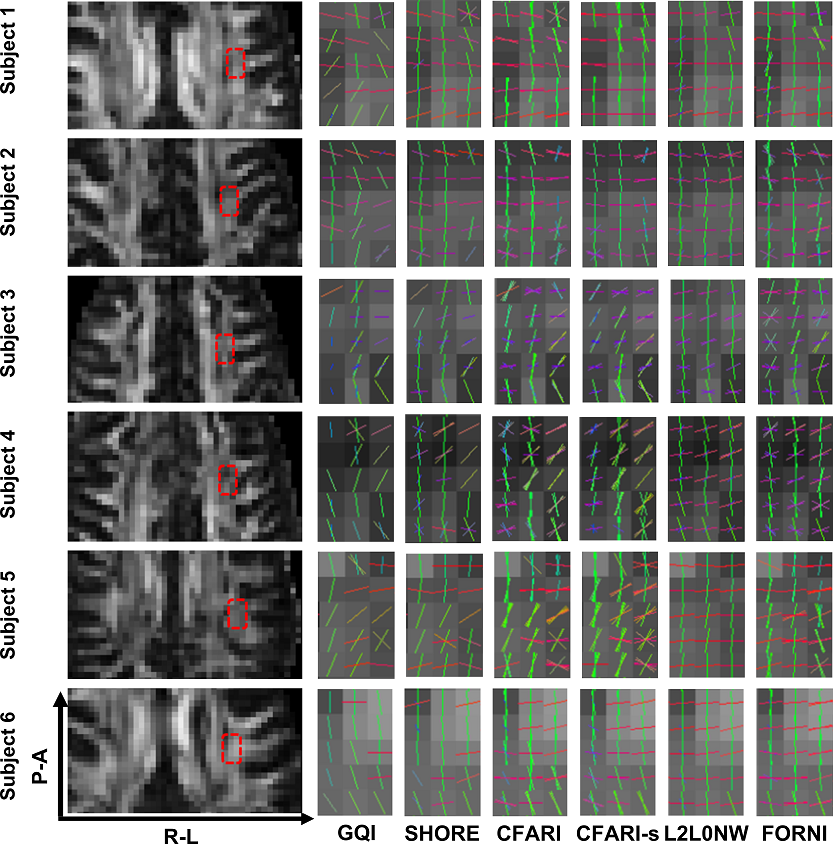}
\caption{FO estimation on brain dMRI (overlaid on the FA map) for all
  six subjects, which is focused on the crossing of SLF and the lateral CC
  (axial view). The FA image and the focused region are shown in the
  left column.}
\label{fig:slf}
\end{figure}

\begin{figure}[!t]
		\centering
		\includegraphics[width=0.99\columnwidth]{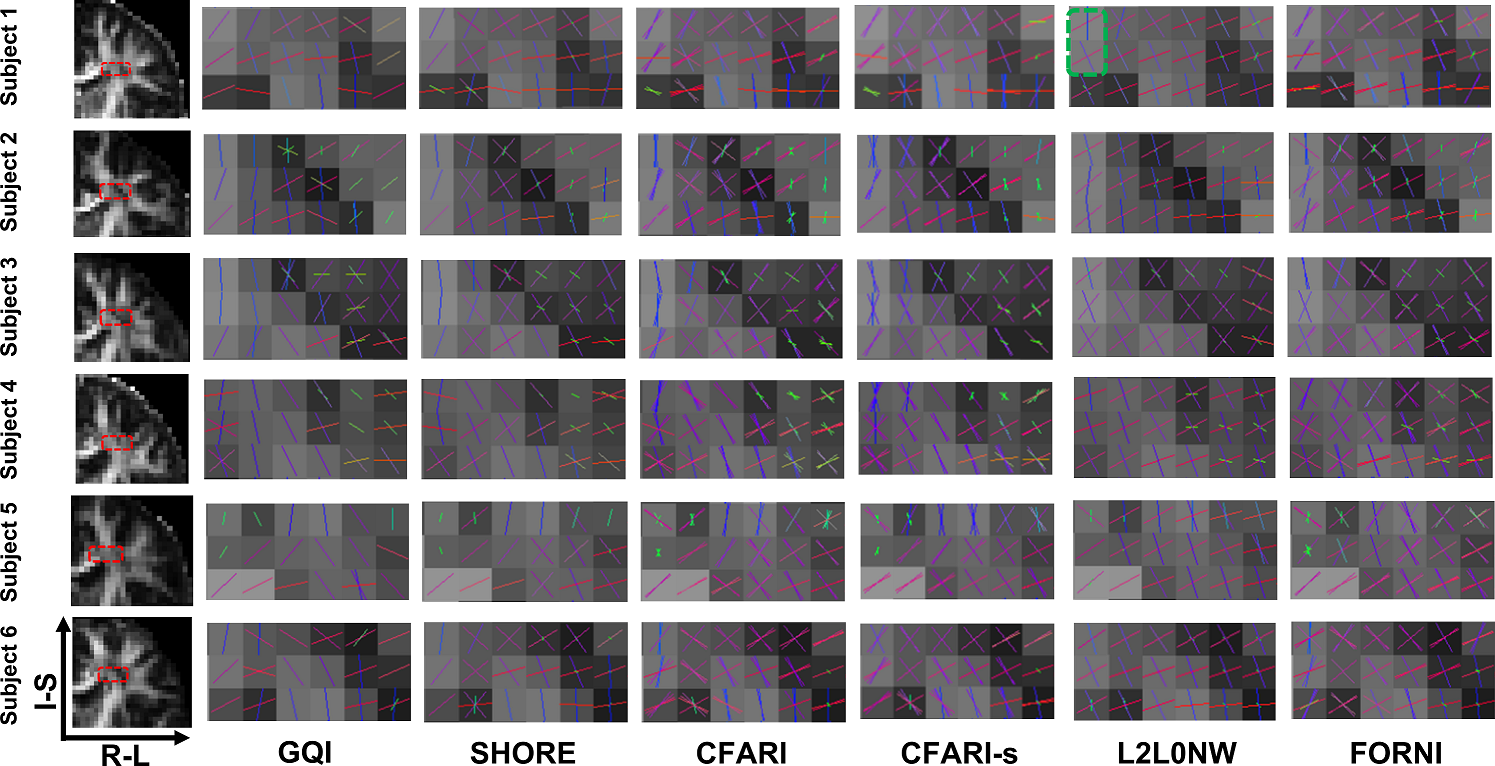}
\caption{FO estimation on brain dMRI (overlaid on the FA map) for all
  six subjects, which is focused on the crossing of the lateral CC and CST
  (coronal view). The FA image and the focused region are shown in the
  left column. Note the region highlighted by the green box where L2L0NW does not produces FOs that correspond to the geometry of the superior CC.}
\label{fig:cc}
\end{figure}

FORNI was applied to the six subjects in an \textit{in vivo}
brain dMRI dataset. The images were acquired on a 3T MRI scanner
(Magnetom Trio, Siemens, Erlangen, Germany). Two $b$-values were used
($b=1000~\mathrm{s}/\mathrm{mm}^{2}$ and
$2000~\mathrm{s}/\mathrm{mm}^{2}$). Each $b$-value is associated with
30 gradient directions and each DWI has two repeated scans. Twelve
$b0$ images were also acquired.   
The resolution is 2.7~mm isotropic (matrix size: $84\times 84$) and
the number of slices is 48. 
The SNR on the $b0$ image is close to 20 in the dMRI dataset.
Motion correction and diffusion tensor estimation were performed by
CATNAP~\citep{CATNAP}. 

FORNI (with $\alpha=0.8$, $\beta=0.5$, and $\mu=3.0$) was compared with GQI~\citep{gqi}, 
SHORE~\citep{merlet2013}, CFARI~\citep{landman}, \mbox{CFARI-s}~\citep{sigurdsson}, and L2L0NW~\citep{auria}.
The eigenvalues of the basis tensors were $\lambda_{1}=2.0\times
10^{-3}~\mathrm{mm}^{2}/\mathrm{s}$ and
$\lambda_{2}=\lambda_{3}=5.0\times
10^{-4}~\mathrm{mm}^{2}/\mathrm{s}$ as suggested by~\citet{landman}. The FORNI processing took around
one hour for each dMRI data. 

\begin{figure}[t]
		\centering
		\includegraphics[width=0.99\columnwidth]{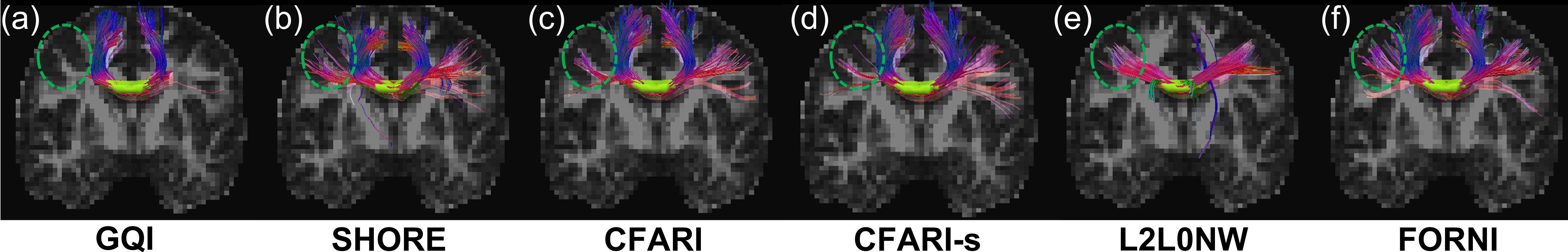}
\caption{A representative result (Subject 1) of fiber tracking using the INFACT
  tracking~\citep{landman} overlaid on the FA map in the coronal
  view. The seeding region is represented as the yellow volume. Note
  the highlighted region where more lateral CC fiber streamlines were
  tracked using FOs computed by FORNI than GQI, SHORE, CFARI, and CFARI-s. The visualization was created in
  TrackVis~\citep{trackvis}.}
\label{fig:infact}
\end{figure}

\begin{figure}[t]
		\centering
		\includegraphics[width=0.99\columnwidth]{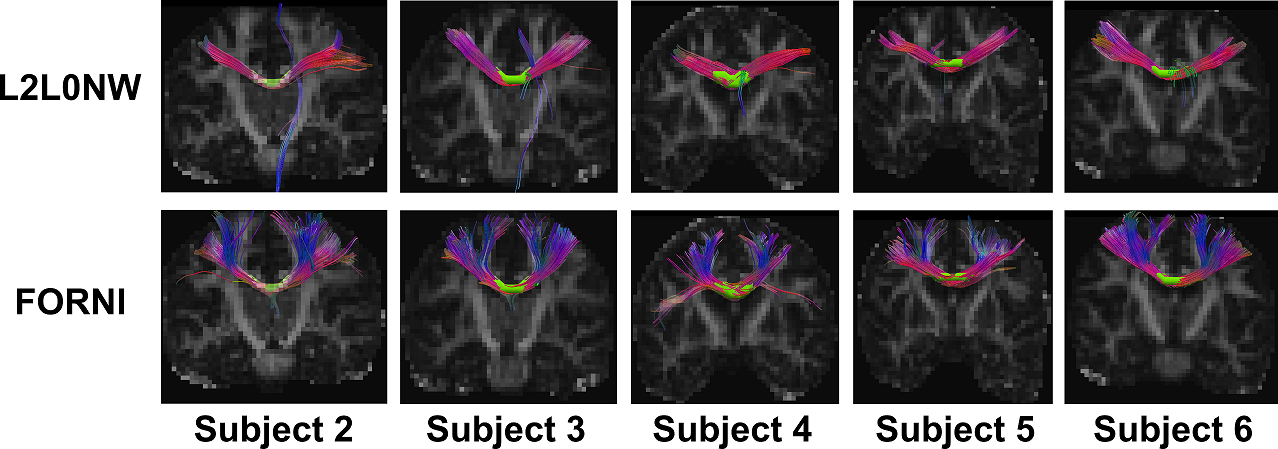}
\caption{Fiber tracking results seeded in CC on Subjects 2--6 using L2L0NW and FORNI FOs. The seeding ROIs are indicated by the yellow volumes. The visualization was created in TrackVis~\citep{trackvis}.}
\label{fig:infact_cc}
\end{figure}

We highlight two regions for evaluation of the results on the six
subjects: the crossing region of the superior longitudinal fasciculus
(SLF) and the lateral corpus callosum (CC) in the axial view
(Figure~\ref{fig:slf}) and the crossing region of the lateral CC and the
corticospinal tract (CST) in the coronal view (Figure~\ref{fig:cc}).
The results are shown with FA images in the left column. 
Compared to GQI, SHORE, CFARI, and CFARI-s, both L2L0NW and FORNI
produces smooth FO estimation results and better identifies the crossing patterns
in all cases. However, we note that in the region containing highly curved parts of the superior CC, for example, the one highlighted by the green box on Subject 1 in Figure~\ref{fig:cc}, L2L0NW does not generate the FOs that correspond to the pathway of the superior CC. The effects of these kinds of errors will be better illustrated below in the fiber tracking results of Figures~\ref{fig:infact} and \ref{fig:infact_cc}. 

To further validate FORNI and compare it with the competing methods we used the INFACT algorithm~\citep{landman}
to carry out fiber tracking on the results of each algorithm. INFACT
is a deterministic streamlining algorithm which extends the FACT
algorithm~\citep{mori} to our case where there are multiple FOs per
voxel. We used an FA threshold of 0.15 and a turning angle threshold of $40^{\circ}$,
which are common settings for tractography~\citep{wahl,glasser,kaplan}. The
seeds were placed in the noncrossing region of CC. 
A representative case (Subject 1) is shown in
Figure~\ref{fig:infact}, where the FORNI result can be 
compared to the results of GQI, SHORE, CFARI, CFARI-s, and L2L0NW.
Here each segment of the fiber streamlines is 
color-coded by the standard DTI color scheme (red: left--right; green:
front--back; and blue: up--down)~\citep{pajevic}. 
It can be seen that the lateral CC fiber streamlines are
tracked better using FOs estimated by FORNI than GQI, SHORE, CFARI, and CFARI-s. In L2L0NW results, although the lateral CC is also well tracked, the superior CC is mostly missing, which is consistent with the FO estimation highlighted by the green box in Figure~\ref{fig:cc}. This missing of the superior CC also exists in the other subjects, which is shown in Figure~\ref{fig:infact_cc}. Here, we applied fiber tracking using FORNI and L2L0NW FOs on the other five subjects with seeds placed in CC. We can see that FORNI is able to produce both superior and lateral CC but L2L0NW misses the superior CC.

\section{Discussion}
\label{sec:discussion}

FORNI uses weighted $\ell_{1}$-norm regularization as
in the FIEBR algorithm~\citep{CMIG}, but the key ideas are very different between the two
algorithms. 
First, the prior directions/likely FOs at each voxel are determined very
differently. In~\citet{CMIG}, the prior directions are calculated based on
the anatomical information of known tracts and it is aimed at
resolving crossing fibers with a very limited number of gradient
directions (around~12). 
Its performance could be affected by prior direction inaccuracies, for example, caused by inaccurate localization of tracts using registration. In the results in Figure~\ref{fig:tongue} we can see that FIEBR can miss the crossing patterns due to the inaccurate specification of anatomical priors.
In the proposed method no anatomical
information is required, and the purpose is to improve FO estimation
by incorporating spatial coherence of FOs. Second, the FOs are
estimated at each voxel independently in~\citet{CMIG} and no
interaction between voxels is encoded while the proposed method
jointly estimates the FOs in all voxels due to the interaction.

The L2L0NW algorithm~\citep{auria} also uses weighted $\ell_{1}$-norm to model the interaction between neighbors to enforce spatial regularization in its objective function. However, the motivation and the actual determination of the weighting are quite different between L2L0NW and FORNI. First, our method was motivated by the framework developed in~\citet{CMIG}, where the weighted $\ell_{1}$-norm is derived in a \textit{maximum a posteriori} framework and is a consequence of modeling the prior density with a Laplace distribution and a term that encourages basis directions close to certain prior directions. The L2L0NW algorithm is motivated by the iterative reweighting scheme that seeks to better solve the $\ell_{0}$-norm minimization problem. 
Second, the weighting is determined differently. L2L0NW directly uses all the neighbor information in the weighting, and a neighbor FO has no influence on directions farther than $15^{\circ}$ and uniform influence on directions within $15^{\circ}$. For a region with a highly curved tract, for example, the turning of the superior CC in~Figure~\ref{fig:cc}, it is possible that the desired FO is more than $15^{\circ}$ away from its neighbor FOs and is not sufficiently encouraged. In FORNI, we process the information in the neighbors and extract likely FOs, and the weight decreases as the directions are closer to likely FOs. This strategy avoids the cutoff effect when a threshold of $15^{\circ}$ is used in~\citet{auria}. In addition, we have used a voxel similarity term to avoid leakage of FOs, where the existence of undesired FOs at a voxel is a result of the impact of the FOs similar to the undesired ones in its neighbors.
The voxel similarity puts higher weights on more similar neighbors and ensures anisotropic FO spatial consistency. This is especially important at tract boundaries and highly curved regions of tracts to suppress the influence of undesired FOs. As seen in the phantom experiment (Figure~\ref{fig:phantom_fo_all}) and the brain dMRI results (Figure~\ref{fig:cc}), L2L0NW can have leaking FOs at tract boundaries and miss the FOs of the curved superior CC, respectively, but this is avoided in FORNI.

A possible limitation of using the tensor distance in the voxel similarity is that at the boundary of the noncrossing part of a tract and its crossing part, the information in the noncrossing neighbors does not influence the current voxel as much as the crossing neighbors. It may be interesting to allow greater influence of noncrossing voxels on the crossing voxels that belong to the same tract while maintaining the avoidance of leaking of the FOs belonging to a different tract from the crossing voxels to noncrossing voxels.

In FORNI, the interaction between neighbors are decoupled using a BCD
strategy. If a different update order of the voxels were used, the
results of each iteration could be different. However, because
multiple iterations are applied, the final results are expected to be
very similar even if the update order is changed. But it would be
interesting to explore adaptive sweeping patterns, such as~\citet{li},
so that the optimization is less dependent on the voxel order.

Instead of enforcing pairwise similarity between neighbor voxels,
FORNI explicitly models neighbor FO information in the FO estimation
by placing different penalties on the basis. Yet it is possible to
combine the pairwise similarity with FORNI. For example, a
straightforward improvement could be adding post-smoothing of FOs,
such as~\citet{sigurdsson}, to the FORNI results.  Indeed, the FORNI
processing and the post-smoothing could be performed alternately with
many iterations. These alternating iterative steps could actually
correspond to the optimization of some unknown form of objective
functions, which can be explored by future work to develop more
powerful FO estimation algorithms.

Besides the directional information of FOs, the microstructural property has also been a quantity of interest computed from dMRI~\citep{zhang,daducci2015,alexander,auria2015}, which quantifies the tissue structure at mesoscale~\citep{reisert2014}, and recent work has further explored joint estimation of the microstructural characteristics and FOs~\citep{girard}. 
Among these works, \citet{daducci2015} and \citet{auria2015} reformulate the estimation of microstructural properties by using a dictionary. Using this reformulation, it is possible to extend our framework to incorporate the estimation of the tissue organization, which could be improved by the incorporation of spatial smoothness. 

At higher $b$-values, the diffusion is not Gaussian due to the restriction effects~\citep{cohen}. Thus, there can be model inaccuracies caused by using the tensor basis. The sparsity and spatial regularization enforced by weighted $\ell_{1}$-norm terms could alleviate the issue. And it is possible to use the extension where microstructural properties are jointly estimated with FOs to account for the slow diffusing components. 

\section{Summary and Conclusion}
\label{sec:conclusion}

In this work, we have proposed FORNI, an FO estimation algorithm that uses
neighborhood information.  A fixed tensor basis is used to
represent the diffusion signals.  To ensure spatial coherence, the
directional information in the neighbors is explicitly modeled in
weighted $\ell_{1}$-norm regularization terms.  The resulting
objective function is optimized using a BCD strategy and a
parallelization approach to speeding up processing is presented.  The
proposed method was applied to a digital crossing phantom, \textit{ex
  vivo} tongue dMRI data, and \textit{in vivo} brain dMRI data. The
results demonstrate that the proposed method is able to use
neighborhood information to improve FO estimation. 

\section*{Acknowledgement}

This work is supported by NIH\slash NINDS 5R01NS056307 and NIH\slash
NINDS 1R21NS082891. 

\appendix
\section{Optimization of the Weighted $\ell_{1}$-norm Regularized Least Squares Problem}
\label{sec:appendix}

In Eq.~(\ref{equ:bcd2}) the diagonal entries of (diagonal matrix) $\textbf{C}_{m}^{t}$
are given by (see Eq.~(\ref{eq:diagonalweights}))
\begin{eqnarray}
C_{m,i}^{t} = \frac{1 - \alpha\max\limits_{p=1,\ldots,U_m^{t}}|\bm{v}_{i}\cdot
  \bm{u}_{m,p}^{t}|}{\min\limits_{q=1,\ldots,N}\left(1 -
    \alpha\max\limits_{p=1,\ldots,U_m^{t}} |\bm{v}_{q}\cdot \bm{u}_{m,p}^{t}|\right)}\,,
\quad i=1,\ldots,N \,. 
\end{eqnarray}
Here, $\bm{u}_{m,p}^{t}$ represents the $p$-th likely FO for voxel $m$ at iteration $t$ and is to be specified, and $U_m^{t}$ is the number of likely FOs for voxel $m$ at iteration $t$.
Eq.~(\ref{equ:bcd2}) explicitly acknowledges the fact that at time
$t$, the estimate of $\bm{f}_{m}$ uses information from voxels
that have already been updated at time $t$ as well as information from
voxels that were updated at the previous time $t-1$. (This is a
Gauss-Seidel rather than a Jacobi update strategy.) Using this fact, the computed aggregate basis-neighbor similarity function for voxel $m$ at time $t$ is 
\begin{eqnarray}
R_{m}^{t}(i) = \sum\limits_{n\in
  \mathcal{N}_{m}}w_{m,n}\max\limits_{j=1,\ldots,W_{n}^{t-\mathds{1}_{n > m}}}|\bm{v}_{i}\cdot
\bm{w}_{n,j}^{t-\mathds{1}_{n > m}}|, 
\end{eqnarray}
where $\mathds{1}$ is an indicator
function providing a shorthand notation to specify whether FOs at time
$t$ or $t-1$ are being used.  
The likely FOs $\mathcal{U}_{m}^{t}$ at time~$t$ are then computed according to Eq.~(\ref{equ:likely_FOs}). 
With these definitions, Eq.~(\ref{equ:bcd2}) is fully specified.

Eq.~(\ref{equ:bcd2}) is a weighted $\ell_{1}$-norm
regularized least squares problem which can be converted to an
$\ell_{1}$-norm regularized least squares problem. First, we define a
new variable 
$\bm{g}_{m}^{t}=\textbf{C}_{m}^{t}\bm{f}_{m}$. Since
$\textbf{C}_{m}^{t}$ is a diagonal matrix and $C_{m,i}^{t}>0$,
$\textbf{C}_{m}^{t}$ is invertible and $\bm{f}_{m} =
(\textbf{C}_{m}^{t})^{-1}\bm{g}_{m}^{t}$. Then, by defining
$\tilde{\textbf{G}}_{m}^{t}=\textbf{G}(\textbf{C}_{m}^{t})^{-1}$, we
have 
\begin{eqnarray}
\hat{\bm{g}}_{m}^{t} = \argmin\limits_{\bm{g}_{m}^{t}\geq
  \bm{0}}||\tilde{\textbf{G}}_{m}^{t}\bm{g}_{m}^{t}-\bm{y}_{m}||_{2}^{2}
+ \beta ||\bm{g}_{m}^{t}||_{1}, 
\label{equ:obj2}
\end{eqnarray}
which we solve using the efficient optimization method for compressed
sensing reported in~\citet{kim}. The mixture fractions can be
estimated as  
\begin{eqnarray}
\hat{\bm{f}}_{m}^{t}=(\textbf{C}_{m}^{t})^{-1}\hat{\bm{g}}_{m}^{t}.
\label{equ:fm}
\end{eqnarray}

\bibliographystyle{model2-names}
\bibliography{refs}

\begin{thebibliography}{69}
\expandafter\ifx\csname natexlab\endcsname\relax\def\natexlab#1{#1}\fi
\providecommand{\url}[1]{\texttt{#1}}
\providecommand{\href}[2]{#2}
\providecommand{\path}[1]{#1}
\providecommand{\DOIprefix}{doi:}
\providecommand{\ArXivprefix}{arXiv:}
\providecommand{\URLprefix}{URL: }
\providecommand{\Pubmedprefix}{pmid:}
\providecommand{\doi}[1]{\href{http://dx.doi.org/#1}{\path{#1}}}
\providecommand{\Pubmed}[1]{\href{pmid:#1}{\path{#1}}}
\providecommand{\bibinfo}[2]{#2}
\ifx\xfnm\relax \def\xfnm[#1]{\unskip,\space#1}\fi
\bibitem[{Alexander et~al.(2010)Alexander, Hubbard, Hall, Moore, Ptito, Parker
  and Dyrby}]{alexander}
\bibinfo{author}{Alexander, D.C.}, \bibinfo{author}{Hubbard, P.L.},
  \bibinfo{author}{Hall, M.G.}, \bibinfo{author}{Moore, E.A.},
  \bibinfo{author}{Ptito, M.}, \bibinfo{author}{Parker, G.J.},
  \bibinfo{author}{Dyrby, T.B.}, \bibinfo{year}{2010}.
\newblock \bibinfo{title}{Orientationally invariant indices of axon diameter
  and density from diffusion {MRI}}.
\newblock \bibinfo{journal}{NeuroImage} \bibinfo{volume}{52},
  \bibinfo{pages}{1374--1389}.
\bibitem[{Aranda et~al.(2014)Aranda, Rivera and Ramirez-Manzanares}]{aranda}
\bibinfo{author}{Aranda, R.}, \bibinfo{author}{Rivera, M.},
  \bibinfo{author}{Ramirez-Manzanares, A.}, \bibinfo{year}{2014}.
\newblock \bibinfo{title}{A flocking based method for brain tractography}.
\newblock \bibinfo{journal}{Medical Image Analysis} \bibinfo{volume}{18},
  \bibinfo{pages}{515--530}.
\bibitem[{Arsigny et~al.(2006)Arsigny, Fillard, Pennec and Ayache}]{arsigny}
\bibinfo{author}{Arsigny, V.}, \bibinfo{author}{Fillard, P.},
  \bibinfo{author}{Pennec, X.}, \bibinfo{author}{Ayache, N.},
  \bibinfo{year}{2006}.
\newblock \bibinfo{title}{Log-{E}uclidean metrics for fast and simple calculus
  on diffusion tensors}.
\newblock \bibinfo{journal}{Magnetic Resonance in Medicine}
  \bibinfo{volume}{56}, \bibinfo{pages}{411--421}.
\bibitem[{Aur{\'\i}a et~al.(2015a)Aur{\'\i}a, Daducci, Thiran and
  Wiaux}]{auria}
\bibinfo{author}{Aur{\'\i}a, A.}, \bibinfo{author}{Daducci, A.},
  \bibinfo{author}{Thiran, J.P.}, \bibinfo{author}{Wiaux, Y.},
  \bibinfo{year}{2015}a.
\newblock \bibinfo{title}{Structured sparsity for spatially coherent fibre
  orientation estimation in diffusion {MRI}}.
\newblock \bibinfo{journal}{NeuroImage} \bibinfo{volume}{115},
  \bibinfo{pages}{245--255}.
\bibitem[{Aur{\'\i}a et~al.(2015b)Aur{\'\i}a, Romascano, Canales-Rodriguez,
  Wiaux, Dirby, Alexander, Thiran and Daducci}]{auria2015}
\bibinfo{author}{Aur{\'\i}a, A.}, \bibinfo{author}{Romascano, D.P.R.},
  \bibinfo{author}{Canales-Rodriguez, E.}, \bibinfo{author}{Wiaux, Y.},
  \bibinfo{author}{Dirby, T.B.}, \bibinfo{author}{Alexander, D.},
  \bibinfo{author}{Thiran, J.P.}, \bibinfo{author}{Daducci, A.},
  \bibinfo{year}{2015}b.
\newblock \bibinfo{title}{Accelerated microstructure imaging via convex
  optimisation for regions with multiple fibres ({AMICOx})}, in:
  \bibinfo{booktitle}{IEEE International Conference on Image Processing 2015},
  \bibinfo{organization}{IEEE}. pp. \bibinfo{pages}{1673--1676}.
\bibitem[{Basser et~al.(1994)Basser, Mattiello and LeBihan}]{basser}
\bibinfo{author}{Basser, P.J.}, \bibinfo{author}{Mattiello, J.},
  \bibinfo{author}{LeBihan, D.}, \bibinfo{year}{1994}.
\newblock \bibinfo{title}{{MR} diffusion tensor spectroscopy and imaging}.
\newblock \bibinfo{journal}{Biophysical Journal} \bibinfo{volume}{66},
  \bibinfo{pages}{259--267}.
\bibitem[{Basser et~al.(2000)Basser, Pajevic, Pierpaoli, Duda and
  Aldroubi}]{basser2}
\bibinfo{author}{Basser, P.J.}, \bibinfo{author}{Pajevic, S.},
  \bibinfo{author}{Pierpaoli, C.}, \bibinfo{author}{Duda, J.},
  \bibinfo{author}{Aldroubi, A.}, \bibinfo{year}{2000}.
\newblock \bibinfo{title}{In vivo fiber tractography using {DT-MRI} data}.
\newblock \bibinfo{journal}{Magnetic Resonance in Medicine}
  \bibinfo{volume}{44}, \bibinfo{pages}{625--632}.
\newblock \URLprefix
  \url{http://dx.doi.org/10.1002/1522-2594(200010)44:4<625::AID-MRM17>3.0.CO;2-O},
  \DOIprefix\doi{10.1002/1522-2594(200010)44:4<625::AID-MRM17>3.0.CO;2-O}.
\bibitem[{Bazin et~al.(2011)Bazin, Ye, Bogovic, Shiee, Reich, Prince and
  Pham}]{bazin}
\bibinfo{author}{Bazin, P.L.}, \bibinfo{author}{Ye, C.},
  \bibinfo{author}{Bogovic, J.A.}, \bibinfo{author}{Shiee, N.},
  \bibinfo{author}{Reich, D.S.}, \bibinfo{author}{Prince, J.L.},
  \bibinfo{author}{Pham, D.L.}, \bibinfo{year}{2011}.
\newblock \bibinfo{title}{Direct segmentation of the major white matter tracts
  in diffusion tensor images}.
\newblock \bibinfo{journal}{NeuroImage} \bibinfo{volume}{58},
  \bibinfo{pages}{458--468}.
\newblock \DOIprefix\doi{10.1016/j.neuroimage.2011.06.020}.
\bibitem[{Becker et~al.(2014)Becker, Tabelow, Mohammadi, Weiskopf and
  Polzehl}]{becker}
\bibinfo{author}{Becker, S.}, \bibinfo{author}{Tabelow, K.},
  \bibinfo{author}{Mohammadi, S.}, \bibinfo{author}{Weiskopf, N.},
  \bibinfo{author}{Polzehl, J.}, \bibinfo{year}{2014}.
\newblock \bibinfo{title}{Adaptive smoothing of multi-shell diffusion weighted
  magnetic resonance data by {msPOAS}}.
\newblock \bibinfo{journal}{NeuroImage} \bibinfo{volume}{95},
  \bibinfo{pages}{90--105}.
\bibitem[{Behrens et~al.(2007)Behrens, Berg, Jbabdi, Rushworth and
  Woolrich}]{behrens}
\bibinfo{author}{Behrens, T.E.J.}, \bibinfo{author}{Berg, H.J.},
  \bibinfo{author}{Jbabdi, S.}, \bibinfo{author}{Rushworth, M.F.S.},
  \bibinfo{author}{Woolrich, M.W.}, \bibinfo{year}{2007}.
\newblock \bibinfo{title}{Probabilistic diffusion tractography with multiple
  fibre orientations: What can we gain?}
\newblock \bibinfo{journal}{NeuroImage} \bibinfo{volume}{34},
  \bibinfo{pages}{144--155}.
\bibitem[{Bertsekas(1999)}]{bertsekas}
\bibinfo{author}{Bertsekas, D.P.}, \bibinfo{year}{1999}.
\newblock \bibinfo{title}{{Nonlinear Programming, Second Edition}}.
\newblock \bibinfo{publisher}{Athena Scientific}.
\bibitem[{Bilgic et~al.(2012)Bilgic, Setsompop, Cohen-Adad, Yendiki, Wald and
  Adalsteinsson}]{bilgic}
\bibinfo{author}{Bilgic, B.}, \bibinfo{author}{Setsompop, K.},
  \bibinfo{author}{Cohen-Adad, J.}, \bibinfo{author}{Yendiki, A.},
  \bibinfo{author}{Wald, L.L.}, \bibinfo{author}{Adalsteinsson, E.},
  \bibinfo{year}{2012}.
\newblock \bibinfo{title}{Accelerated diffusion spectrum imaging with
  compressed sensing using adaptive dictionaries}.
\newblock \bibinfo{journal}{Magnetic Resonance in Medicine}
  \bibinfo{volume}{68}, \bibinfo{pages}{1747--1754}.
\bibitem[{Cheng et~al.(2014)Cheng, Deriche, Jiang, Shen and Yap}]{cheng2}
\bibinfo{author}{Cheng, J.}, \bibinfo{author}{Deriche, R.},
  \bibinfo{author}{Jiang, T.}, \bibinfo{author}{Shen, D.},
  \bibinfo{author}{Yap, P.T.}, \bibinfo{year}{2014}.
\newblock \bibinfo{title}{{Non-Negative Spherical Deconvolution (NNSD) for
  estimation of fiber Orientation Distribution Function in single-/multi-shell
  diffusion MRI}}.
\newblock \bibinfo{journal}{NeuroImage} \bibinfo{volume}{101},
  \bibinfo{pages}{750--764}.
\bibitem[{Cheng et~al.(2011)Cheng, Jiang and Deriche}]{cheng2011}
\bibinfo{author}{Cheng, J.}, \bibinfo{author}{Jiang, T.},
  \bibinfo{author}{Deriche, R.}, \bibinfo{year}{2011}.
\newblock \bibinfo{title}{Theoretical analysis and practical insights on {EAP}
  estimation via a unified {HARDI} framework}, in: \bibinfo{booktitle}{MICCAI
  Workshop on Computational Diffusion MRI (CDMRI)}.
\bibitem[{Cheng et~al.(2006)Cheng, Magnotta, Wu, Nopoulos, Moser, Paulsen,
  Jorge and Andreasen}]{cheng}
\bibinfo{author}{Cheng, P.}, \bibinfo{author}{Magnotta, V.A.},
  \bibinfo{author}{Wu, D.}, \bibinfo{author}{Nopoulos, P.},
  \bibinfo{author}{Moser, D.J.}, \bibinfo{author}{Paulsen, J.},
  \bibinfo{author}{Jorge, R.}, \bibinfo{author}{Andreasen, N.C.},
  \bibinfo{year}{2006}.
\newblock \bibinfo{title}{Evaluation of the {GTRACT} diffusion tensor
  tractography algorithm: a validation and reliability study.}
\newblock \bibinfo{journal}{NeuroImage} \bibinfo{volume}{31},
  \bibinfo{pages}{1075--1085}.
\bibitem[{Cohen and Assaf(2002)}]{cohen}
\bibinfo{author}{Cohen, Y.}, \bibinfo{author}{Assaf, Y.}, \bibinfo{year}{2002}.
\newblock \bibinfo{title}{High b-value q-space analyzed diffusion-weighted {MRS
  and MRI} in neuronal tissues--a technical review}.
\newblock \bibinfo{journal}{NMR in Biomedicine} \bibinfo{volume}{15},
  \bibinfo{pages}{516--542}.
\bibitem[{Daducci et~al.(2015)Daducci, Canales-Rodr{\'\i}guez, Zhang, Dyrby,
  Alexander and Thiran}]{daducci2015}
\bibinfo{author}{Daducci, A.}, \bibinfo{author}{Canales-Rodr{\'\i}guez, E.J.},
  \bibinfo{author}{Zhang, H.}, \bibinfo{author}{Dyrby, T.B.},
  \bibinfo{author}{Alexander, D.C.}, \bibinfo{author}{Thiran, J.P.},
  \bibinfo{year}{2015}.
\newblock \bibinfo{title}{{Accelerated Microstructure Imaging via Convex
  Optimization (AMICO}) from diffusion {MRI} data}.
\newblock \bibinfo{journal}{NeuroImage} \bibinfo{volume}{105},
  \bibinfo{pages}{32--44}.
\bibitem[{Daducci et~al.(2014)Daducci, Van De~Ville, Thiran and
  Wiaux}]{daducci}
\bibinfo{author}{Daducci, A.}, \bibinfo{author}{Van De~Ville, D.},
  \bibinfo{author}{Thiran, J.P.}, \bibinfo{author}{Wiaux, Y.},
  \bibinfo{year}{2014}.
\newblock \bibinfo{title}{{Sparse regularization for fiber ODF reconstruction:
  From the suboptimality of $\ell_2$ and $\ell_1$ priors to $\ell_0$}}.
\newblock \bibinfo{journal}{Medical Image Analysis} \bibinfo{volume}{18},
  \bibinfo{pages}{820--833}.
\bibitem[{Descoteaux et~al.(2007)Descoteaux, Angelino, Fitzgibbons and
  Deriche}]{qbi}
\bibinfo{author}{Descoteaux, M.}, \bibinfo{author}{Angelino, E.},
  \bibinfo{author}{Fitzgibbons, S.}, \bibinfo{author}{Deriche, R.},
  \bibinfo{year}{2007}.
\newblock \bibinfo{title}{Regularized, fast, and robust analytical q-ball
  imaging}.
\newblock \bibinfo{journal}{Magnetic Resonance in Medicine}
  \bibinfo{volume}{58}, \bibinfo{pages}{497--510}.
\bibitem[{Duits and Franken(2011)}]{duits}
\bibinfo{author}{Duits, R.}, \bibinfo{author}{Franken, E.},
  \bibinfo{year}{2011}.
\newblock \bibinfo{title}{Left-invariant diffusions on the space of positions
  and orientations and their application to crossing-preserving smoothing of
  {HARDI} images}.
\newblock \bibinfo{journal}{International Journal of Computer Vision}
  \bibinfo{volume}{92}, \bibinfo{pages}{231--264}.
\bibitem[{Garyfallidis et~al.(2014)Garyfallidis, Brett, Amirbekian, Rokem, Van
  Der~Walt, Descoteaux, Nimmo-Smith and Contributors}]{DIPY}
\bibinfo{author}{Garyfallidis, E.}, \bibinfo{author}{Brett, M.},
  \bibinfo{author}{Amirbekian, B.}, \bibinfo{author}{Rokem, A.},
  \bibinfo{author}{Van Der~Walt, S.}, \bibinfo{author}{Descoteaux, M.},
  \bibinfo{author}{Nimmo-Smith, I.}, \bibinfo{author}{Contributors, D.},
  \bibinfo{year}{2014}.
\newblock \bibinfo{title}{Dipy, a library for the analysis of diffusion {MRI}
  data}.
\newblock \bibinfo{journal}{Frontiers in Neuroinformatics} \bibinfo{volume}{8},
  \bibinfo{pages}{1--17}.
\bibitem[{Girard et~al.(2015)Girard, Fick, Descoteaux, Deriche and
  Wassermann}]{girard}
\bibinfo{author}{Girard, G.}, \bibinfo{author}{Fick, R.},
  \bibinfo{author}{Descoteaux, M.}, \bibinfo{author}{Deriche, R.},
  \bibinfo{author}{Wassermann, D.}, \bibinfo{year}{2015}.
\newblock \bibinfo{title}{Axtract: microstructure-driven tractography based on
  the ensemble average propagator}, in: \bibinfo{booktitle}{Information
  Processing in Medical Imaging}, \bibinfo{organization}{Springer}. pp.
  \bibinfo{pages}{675--686}.
\bibitem[{Glasser and Rilling(2008)}]{glasser}
\bibinfo{author}{Glasser, M.F.}, \bibinfo{author}{Rilling, J.K.},
  \bibinfo{year}{2008}.
\newblock \bibinfo{title}{{DTI} tractography of the human brain's language
  pathways}.
\newblock \bibinfo{journal}{Cerebral Cortex} \bibinfo{volume}{18},
  \bibinfo{pages}{2471--2482}.
\bibitem[{Hess et~al.(2006)Hess, Mukherjee, Han, Xu and Vigneron}]{hess}
\bibinfo{author}{Hess, C.P.}, \bibinfo{author}{Mukherjee, P.},
  \bibinfo{author}{Han, E.T.}, \bibinfo{author}{Xu, D.},
  \bibinfo{author}{Vigneron, D.B.}, \bibinfo{year}{2006}.
\newblock \bibinfo{title}{Q-ball reconstruction of multimodal fiber
  orientations using the spherical harmonic basis}.
\newblock \bibinfo{journal}{Magnetic Resonance in Medicine}
  \bibinfo{volume}{56}, \bibinfo{pages}{104--117}.
\newblock \URLprefix \url{http://dx.doi.org/10.1002/mrm.20931},
  \DOIprefix\doi{10.1002/mrm.20931}.
\bibitem[{Jenkinson et~al.(2012)Jenkinson, Beckmann, Behrens, Woolrich and
  Smith}]{FSL}
\bibinfo{author}{Jenkinson, M.}, \bibinfo{author}{Beckmann, C.F.},
  \bibinfo{author}{Behrens, T.E.J.}, \bibinfo{author}{Woolrich, M.W.},
  \bibinfo{author}{Smith, S.M.}, \bibinfo{year}{2012}.
\newblock \bibinfo{title}{{FSL}}.
\newblock \bibinfo{journal}{NeuroImage} \bibinfo{volume}{62},
  \bibinfo{pages}{782--790}.
\newblock \DOIprefix\doi{10.1016/j.neuroimage.2011.09.015}.
\bibitem[{Jeurissen et~al.(2014)Jeurissen, Tournier, Dhollander, Connelly and
  Sijbers}]{jeurissen}
\bibinfo{author}{Jeurissen, B.}, \bibinfo{author}{Tournier, J.D.},
  \bibinfo{author}{Dhollander, T.}, \bibinfo{author}{Connelly, A.},
  \bibinfo{author}{Sijbers, J.}, \bibinfo{year}{2014}.
\newblock \bibinfo{title}{{Multi-tissue constrained spherical deconvolution for
  improved analysis of multi-shell diffusion MRI data}}.
\newblock \bibinfo{journal}{NeuroImage} \bibinfo{volume}{103},
  \bibinfo{pages}{411--426}.
\bibitem[{Jian and Vemuri(2007)}]{jian}
\bibinfo{author}{Jian, B.}, \bibinfo{author}{Vemuri, B.C.},
  \bibinfo{year}{2007}.
\newblock \bibinfo{title}{A unified computational framework for deconvolution
  to reconstruct multiple fibers from diffusion weighted {MRI}}.
\newblock \bibinfo{journal}{IEEE Transactions on Medical Imaging}
  \bibinfo{volume}{26}, \bibinfo{pages}{1464--1471}.
\bibitem[{Johansen-Berg and Behrens(2013)}]{johansen}
\bibinfo{author}{Johansen-Berg, H.}, \bibinfo{author}{Behrens, T.E.J.},
  \bibinfo{year}{2013}.
\newblock \bibinfo{title}{Diffusion {MRI}: from quantitative measurement to in
  vivo neuroanatomy}.
\newblock \bibinfo{publisher}{Academic Press}.
\bibitem[{Kaplan et~al.(2010)Kaplan, Naeser, Martin, Ho, Wang, Baker and
  Pascual-Leone}]{kaplan}
\bibinfo{author}{Kaplan, E.}, \bibinfo{author}{Naeser, M.A.},
  \bibinfo{author}{Martin, P.I.}, \bibinfo{author}{Ho, M.},
  \bibinfo{author}{Wang, Y.}, \bibinfo{author}{Baker, E.},
  \bibinfo{author}{Pascual-Leone, A.}, \bibinfo{year}{2010}.
\newblock \bibinfo{title}{Horizontal portion of arcuate fasciculus fibers track
  to pars opercularis, not pars triangularis, in right and left hemispheres: a
  {DTI} study}.
\newblock \bibinfo{journal}{NeuroImage} \bibinfo{volume}{52},
  \bibinfo{pages}{436--444}.
\bibitem[{Kim et~al.(2007)Kim, Koh, Lustig and Boyd}]{kim}
\bibinfo{author}{Kim, S.J.}, \bibinfo{author}{Koh, K.},
  \bibinfo{author}{Lustig, M.}, \bibinfo{author}{Boyd, S.},
  \bibinfo{year}{2007}.
\newblock \bibinfo{title}{An efficient method for compressed sensing}, in:
  \bibinfo{booktitle}{IEEE International Conference on Image Processing}, pp.
  \bibinfo{pages}{117--120}.
\bibitem[{Landman et~al.(2012)Landman, Bogovic, Wan, ElShahaby, Bazin and
  Prince}]{landman}
\bibinfo{author}{Landman, B.A.}, \bibinfo{author}{Bogovic, J.A.},
  \bibinfo{author}{Wan, H.}, \bibinfo{author}{ElShahaby, F.E.Z.},
  \bibinfo{author}{Bazin, P.L.}, \bibinfo{author}{Prince, J.L.},
  \bibinfo{year}{2012}.
\newblock \bibinfo{title}{Resolution of crossing fibers with constrained
  compressed sensing using diffusion tensor {MRI}}.
\newblock \bibinfo{journal}{NeuroImage} \bibinfo{volume}{59},
  \bibinfo{pages}{2175--2186}.
\bibitem[{Landman et~al.(2007)Landman, Farrell, Patel, Mori and
  Prince}]{CATNAP}
\bibinfo{author}{Landman, B.A.}, \bibinfo{author}{Farrell, J.A.D.},
  \bibinfo{author}{Patel, N.L.}, \bibinfo{author}{Mori, S.},
  \bibinfo{author}{Prince, J.L.}, \bibinfo{year}{2007}.
\newblock \bibinfo{title}{{DTI} fiber tracking: the importance of adjusting
  {DTI} gradient tables for motion correction. {CATNAP} {-} a tool to simplify
  and accelerate {DTI} analysis}, in: \bibinfo{booktitle}{Proc. Org Human Brain
  Mapping 13th Annual Meeting}.
\bibitem[{Li and Osher(2009)}]{li}
\bibinfo{author}{Li, Y.}, \bibinfo{author}{Osher, S.}, \bibinfo{year}{2009}.
\newblock \bibinfo{title}{Coordinate descent optimization for $\ell^{1}$
  minimization with application to compressed sensing; a greedy algorithm}.
\newblock \bibinfo{journal}{Inverse Problems and Imaging} \bibinfo{volume}{3},
  \bibinfo{pages}{487--503}.
\bibitem[{Liu et~al.(2015)Liu, Yuan, Guo and Xu}]{liu}
\bibinfo{author}{Liu, X.}, \bibinfo{author}{Yuan, Z.}, \bibinfo{author}{Guo,
  Z.}, \bibinfo{author}{Xu, D.}, \bibinfo{year}{2015}.
\newblock \bibinfo{title}{{A localized Richardson--Lucy algorithm for fiber
  orientation estimation in high angular resolution diffusion imaging}}.
\newblock \bibinfo{journal}{Medical Physics} \bibinfo{volume}{42},
  \bibinfo{pages}{2524--2539}.
\bibitem[{Lucas et~al.(2010)Lucas, Bogovic, Carass, Bazin, Prince, Pham and
  Landman}]{lucas}
\bibinfo{author}{Lucas, B.C.}, \bibinfo{author}{Bogovic, J.A.},
  \bibinfo{author}{Carass, A.}, \bibinfo{author}{Bazin, P.L.},
  \bibinfo{author}{Prince, J.L.}, \bibinfo{author}{Pham, D.L.},
  \bibinfo{author}{Landman, B.A.}, \bibinfo{year}{2010}.
\newblock \bibinfo{title}{The {Java} image science toolkit {(JIST)} for rapid
  prototyping and publishing of neuroimaging software}.
\newblock \bibinfo{journal}{Neuroinformatics} \bibinfo{volume}{8},
  \bibinfo{pages}{5--17}.
\bibitem[{Merlet et~al.(2012)Merlet, Caruyer and Deriche}]{merlet}
\bibinfo{author}{Merlet, S.}, \bibinfo{author}{Caruyer, E.},
  \bibinfo{author}{Deriche, R.}, \bibinfo{year}{2012}.
\newblock \bibinfo{title}{Parametric dictionary learning for modeling eap and
  odf in diffusion {MRI}}, in: \bibinfo{booktitle}{Medical Image Computing and
  Computer-Assisted Intervention--MICCAI 2012}. \bibinfo{publisher}{Springer},
  pp. \bibinfo{pages}{10--17}.
\bibitem[{Merlet and Deriche(2013)}]{merlet2013}
\bibinfo{author}{Merlet, S.L.}, \bibinfo{author}{Deriche, R.},
  \bibinfo{year}{2013}.
\newblock \bibinfo{title}{{Continuous diffusion signal, EAP and ODF estimation
  via Compressive Sensing in diffusion MRI}}.
\newblock \bibinfo{journal}{Medical Image Analysis} \bibinfo{volume}{17},
  \bibinfo{pages}{556--572}.
\bibitem[{Michailovich et~al.(2011)Michailovich, Rathi and
  Dolui}]{michailovich}
\bibinfo{author}{Michailovich, O.}, \bibinfo{author}{Rathi, Y.},
  \bibinfo{author}{Dolui, S.}, \bibinfo{year}{2011}.
\newblock \bibinfo{title}{Spatially regularized compressed sensing for high
  angular resolution diffusion imaging}.
\newblock \bibinfo{journal}{IEEE Transactions on Medical Imaging}
  \bibinfo{volume}{30}, \bibinfo{pages}{1100--1115}.
\bibitem[{Mori et~al.(1999)Mori, Crain, Chacko and Van~Zijl}]{mori}
\bibinfo{author}{Mori, S.}, \bibinfo{author}{Crain, B.J.},
  \bibinfo{author}{Chacko, V.P.}, \bibinfo{author}{Van~Zijl, P.},
  \bibinfo{year}{1999}.
\newblock \bibinfo{title}{Three-dimensional tracking of axonal projections in
  the brain by magnetic resonance imaging}.
\newblock \bibinfo{journal}{Annals of Neurology} \bibinfo{volume}{45},
  \bibinfo{pages}{265--269}.
\bibitem[{Nazem-Zadeh et~al.(2011)Nazem-Zadeh, Davoodi-Bojd and
  Soltanian-Zadeh}]{nazem}
\bibinfo{author}{Nazem-Zadeh, M.R.}, \bibinfo{author}{Davoodi-Bojd, E.},
  \bibinfo{author}{Soltanian-Zadeh, H.}, \bibinfo{year}{2011}.
\newblock \bibinfo{title}{Atlas-based fiber bundle segmentation using principal
  diffusion directions and spherical harmonic coefficients}.
\newblock \bibinfo{journal}{NeuroImage} \bibinfo{volume}{54},
  \bibinfo{pages}{S146--S164}.
\bibitem[{{\"O}zarslan et~al.(2006){\"O}zarslan, Shepherd, Vemuri, Blackband
  and Mareci}]{ozarslan}
\bibinfo{author}{{\"O}zarslan, E.}, \bibinfo{author}{Shepherd, T.M.},
  \bibinfo{author}{Vemuri, B.C.}, \bibinfo{author}{Blackband, S.J.},
  \bibinfo{author}{Mareci, T.H.}, \bibinfo{year}{2006}.
\newblock \bibinfo{title}{Resolution of complex tissue microarchitecture using
  the diffusion orientation transform ({DOT})}.
\newblock \bibinfo{journal}{NeuroImage} \bibinfo{volume}{31},
  \bibinfo{pages}{1086--1103}.
\bibitem[{Pajevic and Pierpaoli(1999)}]{pajevic}
\bibinfo{author}{Pajevic, S.}, \bibinfo{author}{Pierpaoli, C.},
  \bibinfo{year}{1999}.
\newblock \bibinfo{title}{Color schemes to represent the orientation of
  anisotropic tissues from diffusion tensor data: application to white matter
  fiber tract mapping in the human brain}.
\newblock \bibinfo{journal}{Magnetic Resonance in Medicine}
  \bibinfo{volume}{42}, \bibinfo{pages}{526--540}.
\bibitem[{Peled et~al.(2006)Peled, Friman, Jolesz and Westin}]{peled}
\bibinfo{author}{Peled, S.}, \bibinfo{author}{Friman, O.},
  \bibinfo{author}{Jolesz, F.}, \bibinfo{author}{Westin, C.F.},
  \bibinfo{year}{2006}.
\newblock \bibinfo{title}{Geometrically constrained two-tensor model for
  crossing tracts in {DWI}}.
\newblock \bibinfo{journal}{Magnetic Resonance Imaging} \bibinfo{volume}{24},
  \bibinfo{pages}{1263--1270}.
\bibitem[{Pickalov and Basser(2006)}]{pickalov}
\bibinfo{author}{Pickalov, V.}, \bibinfo{author}{Basser, P.J.},
  \bibinfo{year}{2006}.
\newblock \bibinfo{title}{3-{D} tomographic reconstruction of the average
  propagator from {MRI} data}, in: \bibinfo{booktitle}{3rd IEEE International
  Symposium on Biomedical Imaging: Nano to Macro},
  \bibinfo{organization}{IEEE}. pp. \bibinfo{pages}{710--713}.
\bibitem[{Qazi et~al.(2009)Qazi, Radmanesh, O'Donnell, Kindlmann, Peled,
  Whalen, Westin and Golby}]{qazi}
\bibinfo{author}{Qazi, A.A.}, \bibinfo{author}{Radmanesh, A.},
  \bibinfo{author}{O'Donnell, L.}, \bibinfo{author}{Kindlmann, G.},
  \bibinfo{author}{Peled, S.}, \bibinfo{author}{Whalen, S.},
  \bibinfo{author}{Westin, C.F.}, \bibinfo{author}{Golby, A.J.},
  \bibinfo{year}{2009}.
\newblock \bibinfo{title}{Resolving crossings in the corticospinal tract by
  two-tensor streamline tractography: Method and clinical assessment using
  f{MRI}}.
\newblock \bibinfo{journal}{NeuroImage} \bibinfo{volume}{47},
  \bibinfo{pages}{98--106}.
\bibitem[{Ramirez-Manzanares et~al.(2007)Ramirez-Manzanares, Rivera, Vemuri,
  Carney and Mareci}]{ramirez}
\bibinfo{author}{Ramirez-Manzanares, A.}, \bibinfo{author}{Rivera, M.},
  \bibinfo{author}{Vemuri, B.C.}, \bibinfo{author}{Carney, P.},
  \bibinfo{author}{Mareci, T.}, \bibinfo{year}{2007}.
\newblock \bibinfo{title}{Diffusion basis functions decomposition for
  estimating white matter intravoxel fiber geometry}.
\newblock \bibinfo{journal}{IEEE Transactions on Medical Imaging}
  \bibinfo{volume}{26}, \bibinfo{pages}{1091--1102}.
\bibitem[{Rathi et~al.(2014)Rathi, Michailovich, Laun, Setsompop, Grant and
  Westin}]{rathi}
\bibinfo{author}{Rathi, Y.}, \bibinfo{author}{Michailovich, O.},
  \bibinfo{author}{Laun, F.}, \bibinfo{author}{Setsompop, K.},
  \bibinfo{author}{Grant, P.E.}, \bibinfo{author}{Westin, C.F.},
  \bibinfo{year}{2014}.
\newblock \bibinfo{title}{Multi-shell diffusion signal recovery from sparse
  measurements}.
\newblock \bibinfo{journal}{Medical Image Analysis} \bibinfo{volume}{18},
  \bibinfo{pages}{1143--1156}.
\bibitem[{Reisert et~al.(2014)Reisert, Kiselev, Dihtal, Kellner and
  Novikov}]{reisert2014}
\bibinfo{author}{Reisert, M.}, \bibinfo{author}{Kiselev, V.},
  \bibinfo{author}{Dihtal, B.}, \bibinfo{author}{Kellner, E.},
  \bibinfo{author}{Novikov, D.}, \bibinfo{year}{2014}.
\newblock \bibinfo{title}{{MesoFT}: Unifying diffusion modelling and fiber
  tracking}, in: \bibinfo{booktitle}{Medical Image Computing and
  Computer-Assisted Intervention--MICCAI 2014}. \bibinfo{publisher}{Springer},
  pp. \bibinfo{pages}{201--208}.
\bibitem[{Reisert and Kiselev(2011)}]{reisert2011}
\bibinfo{author}{Reisert, M.}, \bibinfo{author}{Kiselev, V.G.},
  \bibinfo{year}{2011}.
\newblock \bibinfo{title}{Fiber continuity: An anisotropic prior for {ODF}
  estimation}.
\newblock \bibinfo{journal}{IEEE Transactions on Medical Imaging}
  \bibinfo{volume}{30}, \bibinfo{pages}{1274--1283}.
\bibitem[{Reisert et~al.(2011)Reisert, Mader, Anastasopoulos, Weigel, Schnell
  and Kiselev}]{reisert}
\bibinfo{author}{Reisert, M.}, \bibinfo{author}{Mader, I.},
  \bibinfo{author}{Anastasopoulos, C.}, \bibinfo{author}{Weigel, M.},
  \bibinfo{author}{Schnell, S.}, \bibinfo{author}{Kiselev, V.},
  \bibinfo{year}{2011}.
\newblock \bibinfo{title}{Global fiber reconstruction becomes practical}.
\newblock \bibinfo{journal}{NeuroImage} \bibinfo{volume}{54},
  \bibinfo{pages}{955--962}.
\bibitem[{Sigurdsson and Prince(2014)}]{sigurdsson}
\bibinfo{author}{Sigurdsson, G.A.}, \bibinfo{author}{Prince, J.L.},
  \bibinfo{year}{2014}.
\newblock \bibinfo{title}{Smoothing fields of weighted collections with
  applications to diffusion {MRI} processing}, in: \bibinfo{booktitle}{SPIE
  Medical Imaging}, pp. \bibinfo{pages}{90342D--90342D}.
\bibitem[{Stejskal and Tanner(1965)}]{stejskal}
\bibinfo{author}{Stejskal, E.O.}, \bibinfo{author}{Tanner, J.E.},
  \bibinfo{year}{1965}.
\newblock \bibinfo{title}{Spin diffusion measurements: spin echoes in the
  presence of a time-dependent field gradient}.
\newblock \bibinfo{journal}{The Journal of Chemical Physics}
  \bibinfo{volume}{42}, \bibinfo{pages}{288}.
\bibitem[{Tournier et~al.(2013)Tournier, Calamante and Connelly}]{tournier2013}
\bibinfo{author}{Tournier, J.}, \bibinfo{author}{Calamante, F.},
  \bibinfo{author}{Connelly, A.}, \bibinfo{year}{2013}.
\newblock \bibinfo{title}{A robust spherical deconvolution method for the
  analysis of low {SNR} or low angular resolution diffusion data}, in:
  \bibinfo{booktitle}{International Society for Magnetic Resonance in
  Medicine}, p. \bibinfo{pages}{0772}.
\bibitem[{Tournier et~al.(2007)Tournier, Calamante and Connelly}]{tournier2}
\bibinfo{author}{Tournier, J.D.}, \bibinfo{author}{Calamante, F.},
  \bibinfo{author}{Connelly, A.}, \bibinfo{year}{2007}.
\newblock \bibinfo{title}{Robust determination of the fibre orientation
  distribution in diffusion {MRI}: Non-negativity constrained super-resolved
  spherical deconvolution}.
\newblock \bibinfo{journal}{NeuroImage} \bibinfo{volume}{35},
  \bibinfo{pages}{1459--1472}.
\bibitem[{Tournier et~al.(2004)Tournier, Calamante, Gadian and
  Connelly}]{tournier}
\bibinfo{author}{Tournier, J.D.}, \bibinfo{author}{Calamante, F.},
  \bibinfo{author}{Gadian, D.G.}, \bibinfo{author}{Connelly, A.},
  \bibinfo{year}{2004}.
\newblock \bibinfo{title}{Direct estimation of the fiber orientation density
  function from diffusion-weighted {MRI} data using spherical deconvolution}.
\newblock \bibinfo{journal}{NeuroImage} \bibinfo{volume}{23},
  \bibinfo{pages}{1176--1185}.
\bibitem[{Tuch(2004)}]{qball}
\bibinfo{author}{Tuch, D.S.}, \bibinfo{year}{2004}.
\newblock \bibinfo{title}{Q-ball imaging}.
\newblock \bibinfo{journal}{Magnetic Resonance in Medicine}
  \bibinfo{volume}{52}, \bibinfo{pages}{1358--1372}.
\newblock \URLprefix \url{http://dx.doi.org/10.1002/mrm.20279},
  \DOIprefix\doi{10.1002/mrm.20279}.
\bibitem[{Tuch et~al.(2002)Tuch, Reese, Wiegell, Makris, Belliveau and
  Wedeen}]{HARDI}
\bibinfo{author}{Tuch, D.S.}, \bibinfo{author}{Reese, T.G.},
  \bibinfo{author}{Wiegell, M.R.}, \bibinfo{author}{Makris, N.},
  \bibinfo{author}{Belliveau, J.W.}, \bibinfo{author}{Wedeen, V.J.},
  \bibinfo{year}{2002}.
\newblock \bibinfo{title}{High angular resolution diffusion imaging reveals
  intravoxel white matter fiber heterogeneity}.
\newblock \bibinfo{journal}{Magnetic Resonance in Medicine}
  \bibinfo{volume}{48}, \bibinfo{pages}{577--582}.
\newblock \URLprefix \url{http://dx.doi.org/10.1002/mrm.10268},
  \DOIprefix\doi{10.1002/mrm.10268}.
\bibitem[{Wahl et~al.(2007)Wahl, Lauterbach-Soon, Hattingen, Jung, Singer,
  Volz, Klein, Steinmetz and Ziemann}]{wahl}
\bibinfo{author}{Wahl, M.}, \bibinfo{author}{Lauterbach-Soon, B.},
  \bibinfo{author}{Hattingen, E.}, \bibinfo{author}{Jung, P.},
  \bibinfo{author}{Singer, O.}, \bibinfo{author}{Volz, S.},
  \bibinfo{author}{Klein, J.C.}, \bibinfo{author}{Steinmetz, H.},
  \bibinfo{author}{Ziemann, U.}, \bibinfo{year}{2007}.
\newblock \bibinfo{title}{Human motor corpus callosum: topography, somatotopy,
  and link between microstructure and function}.
\newblock \bibinfo{journal}{The Journal of Neuroscience} \bibinfo{volume}{27},
  \bibinfo{pages}{12132--12138}.
\bibitem[{Wang et~al.(2007)Wang, Benner, Sorensen and Wedeen}]{trackvis}
\bibinfo{author}{Wang, R.}, \bibinfo{author}{Benner, T.},
  \bibinfo{author}{Sorensen, A.G.}, \bibinfo{author}{Wedeen, V.J.},
  \bibinfo{year}{2007}.
\newblock \bibinfo{title}{Diffusion toolkit: a software package for diffusion
  imaging data processing and tractography}, in: \bibinfo{booktitle}{Proc Intl
  Soc Mag Reson Med}, p. \bibinfo{pages}{3720}.
\bibitem[{Wedeen et~al.(2005)Wedeen, Hagmann, Tseng, Reese and Weisskoff}]{DSI}
\bibinfo{author}{Wedeen, V.J.}, \bibinfo{author}{Hagmann, P.},
  \bibinfo{author}{Tseng, W.Y.I.}, \bibinfo{author}{Reese, T.G.},
  \bibinfo{author}{Weisskoff, R.M.}, \bibinfo{year}{2005}.
\newblock \bibinfo{title}{Mapping complex tissue architecture with diffusion
  spectrum magnetic resonance imaging}.
\newblock \bibinfo{journal}{Magnetic Resonance in Medicine}
  \bibinfo{volume}{54}, \bibinfo{pages}{1377--1386}.
\newblock \URLprefix \url{http://dx.doi.org/10.1002/mrm.20642},
  \DOIprefix\doi{10.1002/mrm.20642}.
\bibitem[{Wedeen et~al.(2008)Wedeen, Wang, Schmahmann, Benner, Tseng, Dai,
  Pandya, Hagmann, D'Arceuil and de~Crespigny}]{wedeen2008}
\bibinfo{author}{Wedeen, V.J.}, \bibinfo{author}{Wang, R.},
  \bibinfo{author}{Schmahmann, J.D.}, \bibinfo{author}{Benner, T.},
  \bibinfo{author}{Tseng, W.}, \bibinfo{author}{Dai, G.},
  \bibinfo{author}{Pandya, D.}, \bibinfo{author}{Hagmann, P.},
  \bibinfo{author}{D'Arceuil, H.}, \bibinfo{author}{de~Crespigny, A.J.},
  \bibinfo{year}{2008}.
\newblock \bibinfo{title}{Diffusion spectrum magnetic resonance imaging ({DSI})
  tractography of crossing fibers}.
\newblock \bibinfo{journal}{NeuroImage} \bibinfo{volume}{41},
  \bibinfo{pages}{1267--1277}.
\bibitem[{Ye et~al.(2014)Ye, Carass, Murano, Stone and Prince}]{BAMBI}
\bibinfo{author}{Ye, C.}, \bibinfo{author}{Carass, A.},
  \bibinfo{author}{Murano, E.}, \bibinfo{author}{Stone, M.},
  \bibinfo{author}{Prince, J.L.}, \bibinfo{year}{2014}.
\newblock \bibinfo{title}{A {B}ayesian approach to distinguishing
  interdigitated muscles in the tongue from limited diffusion weighted
  imaging}, in: \bibinfo{booktitle}{Bayesian and grAphical Models for
  Biomedical Imaging}. \bibinfo{publisher}{Springer}. volume
  \bibinfo{volume}{8677} of \textit{\bibinfo{series}{Lecture Notes in Computer
  Science}}, pp. \bibinfo{pages}{13--24}.
\bibitem[{Ye et~al.(2015a)Ye, Murano, Stone and Prince}]{CMIG}
\bibinfo{author}{Ye, C.}, \bibinfo{author}{Murano, E.}, \bibinfo{author}{Stone,
  M.}, \bibinfo{author}{Prince, J.L.}, \bibinfo{year}{2015}a.
\newblock \bibinfo{title}{{A Bayesian approach to distinguishing interdigitated
  tongue muscles from limited diffusion magnetic resonance imaging}}.
\newblock \bibinfo{journal}{Computerized Medical Imaging and Graphics}
  \bibinfo{volume}{45}, \bibinfo{pages}{63--74}.
\newblock \URLprefix
  \url{http://www.sciencedirect.com/science/article/pii/S0895611115001032},
  \DOIprefix\doi{http://dx.doi.org/10.1016/j.compmedimag.2015.07.005}.
\bibitem[{Ye et~al.(2015b)Ye, Yang, Ying and Prince}]{NEIN}
\bibinfo{author}{Ye, C.}, \bibinfo{author}{Yang, Z.}, \bibinfo{author}{Ying,
  S.H.}, \bibinfo{author}{Prince, J.L.}, \bibinfo{year}{2015}b.
\newblock \bibinfo{title}{Segmentation of the cerebellar peduncles using a
  random forest classifier and a multi-object geometric deformable model:
  Application to spinocerebellar ataxia type 6}.
\newblock \bibinfo{journal}{Neuroinformatics} \bibinfo{volume}{13},
  \bibinfo{pages}{367--381}.
\newblock \URLprefix \url{http://dx.doi.org/10.1007/s12021-015-9264-7},
  \DOIprefix\doi{10.1007/s12021-015-9264-7}.
\bibitem[{Ye et~al.(2016)Ye, Zhuo, Gullapalli and Prince}]{CDMRI}
\bibinfo{author}{Ye, C.}, \bibinfo{author}{Zhuo, J.},
  \bibinfo{author}{Gullapalli, R.P.}, \bibinfo{author}{Prince, J.L.},
  \bibinfo{year}{2016}.
\newblock \bibinfo{title}{Estimation of fiber orientations using neighborhood
  information}, in: \bibinfo{booktitle}{{Computational Diffusion MRI: MICCAI
  Workshop, Munich, Germany, October 9th, 2015}}.
  \bibinfo{publisher}{Springer}, pp. \bibinfo{pages}{87--96}.
\newblock \DOIprefix\doi{10.1007/978-3-319-28588-7_8}.
\bibitem[{Yeh et~al.(2010)Yeh, Wedeen and Tseng}]{gqi}
\bibinfo{author}{Yeh, F.C.}, \bibinfo{author}{Wedeen, V.},
  \bibinfo{author}{Tseng, W.Y.}, \bibinfo{year}{2010}.
\newblock \bibinfo{title}{Generalized $q$-sampling imaging}.
\newblock \bibinfo{journal}{IEEE Transactions on Medical Imaging}
  \bibinfo{volume}{29}, \bibinfo{pages}{1626--1635}.
\newblock \DOIprefix\doi{10.1109/TMI.2010.2045126}.
\bibitem[{Yendiki et~al.(2011)Yendiki, Panneck, Srinivasan, Stevens,
  Z{\"o}llei, Augustinack, Wang, Salat, Ehrlich, Behrens, Jbabdi, Gollub and
  Fischl}]{yendiki}
\bibinfo{author}{Yendiki, A.}, \bibinfo{author}{Panneck, P.},
  \bibinfo{author}{Srinivasan, P.}, \bibinfo{author}{Stevens, A.},
  \bibinfo{author}{Z{\"o}llei, L.}, \bibinfo{author}{Augustinack, J.},
  \bibinfo{author}{Wang, R.}, \bibinfo{author}{Salat, D.},
  \bibinfo{author}{Ehrlich, S.}, \bibinfo{author}{Behrens, T.},
  \bibinfo{author}{Jbabdi, S.}, \bibinfo{author}{Gollub, R.},
  \bibinfo{author}{Fischl, B.}, \bibinfo{year}{2011}.
\newblock \bibinfo{title}{Automated probabilistic reconstruction of
  white-matter pathways in health and disease using an atlas of the underlying
  anatomy}.
\newblock \bibinfo{journal}{Frontiers in Neuroinformatics} \bibinfo{volume}{5},
  \bibinfo{pages}{12--23}.
\bibitem[{Zhang et~al.(2012)Zhang, Schneider, Wheeler-Kingshott and
  Alexander}]{zhang}
\bibinfo{author}{Zhang, H.}, \bibinfo{author}{Schneider, T.},
  \bibinfo{author}{Wheeler-Kingshott, C.A.}, \bibinfo{author}{Alexander, D.C.},
  \bibinfo{year}{2012}.
\newblock \bibinfo{title}{{NODDI}: practical in vivo neurite orientation
  dispersion and density imaging of the human brain}.
\newblock \bibinfo{journal}{NeuroImage} \bibinfo{volume}{61},
  \bibinfo{pages}{1000--1016}.
\bibitem[{Zhou et~al.(2014)Zhou, Michailovich and Rathi}]{zhou}
\bibinfo{author}{Zhou, Q.}, \bibinfo{author}{Michailovich, O.},
  \bibinfo{author}{Rathi, Y.}, \bibinfo{year}{2014}.
\newblock \bibinfo{title}{Resolving complex fibre architecture by means of
  sparse spherical deconvolution in the presence of isotropic diffusion}, in:
  \bibinfo{booktitle}{SPIE Medical Imaging}, pp.
  \bibinfo{pages}{903425--903425}.

\end{thebibliography}

\end{document}